\documentclass[11pt]{article}

\usepackage[final]{acl}

\usepackage{times}
\usepackage{latexsym}
\usepackage[T1]{fontenc}
\usepackage[utf8]{inputenc}
\usepackage{microtype}
\usepackage{inconsolata}

\usepackage{amsfonts}
\usepackage{nicefrac}
\usepackage{pifont}       
\usepackage{scalerel}     
\usepackage{fontawesome5} 

\usepackage{graphicx}
\usepackage{subfigure}    
\usepackage{booktabs}     
\usepackage{tabularx}
\usepackage{multirow}
\usepackage{makecell}
\usepackage{array}
\usepackage{colortbl}     

\usepackage[normalem]{ulem} 
\usepackage{enumitem}       
\usepackage{adjustbox}
\usepackage{ragged2e}
\usepackage[edges]{forest}  
\usepackage[framemethod=TikZ]{mdframed}

\usepackage{xcolor}
\usepackage[most]{tcolorbox}

\definecolor{main-blue}{RGB}{56, 107, 187} 
\definecolor{hanred}{RGB}{165,30,50}
\definecolor{hangreen}{RGB}{131,179,54}
\definecolor{hanblue}{RGB}{60,127,198}
\definecolor{hanyellow}{RGB}{247,183,42}
\definecolor{sudisblue}{RGB}{93, 177, 251}
\definecolor{sudisred}{RGB}{252,106,108}

\newtcolorbox{takeaway}[1][]{
    enhanced,
    boxrule=0pt,frame hidden, 
    borderline west={4pt}{0pt}{main-blue}, 
    colback=main-blue!10, 
    coltitle=main-blue, 
    coltext=main-blue,  
    fonttitle=\bfseries\sffamily, 
    fontupper=\small\sffamily,
    attach boxed title to top left={yshift=-10pt,xshift=10pt}, 
    before upper={\setlength{\baselineskip}{1.2\baselineskip}}, 
    boxed title style={boxrule=0pt,frame hidden,colback=main-blue!10}, 
    title={\faLightbulb\ Takeaway}, 
    #1
}

\newtcolorbox{summarybox}[1][]{
    enhanced,
    colback=white, 
    colframe=main-blue, 
    fonttitle=\bfseries\sffamily\Large,
    coltitle=white,
    title={Key Takeaway}, 
    sharp corners, 
    drop shadow, 
    #1
}

\newtcolorbox{minimalbox}[1][]{
    enhanced,
    frame hidden, 
    colback=white, 
    borderline north={1pt}{0pt}{main-blue}, 
    borderline south={1pt}{0pt}{main-blue}, 
    fonttitle=\bfseries\sffamily\color{main-blue},
    title={Takeaway:},
    attach boxed title to top left={yshift=-10pt},
    boxed title style={frame hidden, colback=white},
    left=0pt, right=0pt, 
    #1
}

\newcommand{\MySummaryBox}[1]{%
\begin{tcolorbox}[colback=gray!60, colframe=gray!60, arc=2mm, boxrule=0.5pt,
  left=1mm, right=0mm, top=0mm, bottom=0mm]
\textit{#1}
\end{tcolorbox}%
}

\hypersetup{
    breaklinks,
    citecolor=sudisred,
    colorlinks=true,
    linkcolor=sudisred,
    urlcolor=sudisblue
}

\usepackage{cleveref}
\crefname{appendix}{Appendix}{Appendices}
\Crefname{appendix}{Appendix}{Appendices}

\title{Efficient Inference for Large Vision-Language Models: \\ Bottlenecks, Techniques, and Prospects}

\author{
  \textbf{Jun Zhang\textsuperscript{1,2\ding{44}}},
  \textbf{Yicheng Ji\textsuperscript{1,2\ding{44}}},  
  \textbf{Feiyang Ren\textsuperscript{1,2\ding{44}}}, 
  \textbf{Yihang Li\textsuperscript{1,2\ding{44}}},\\
  \textbf{Bowen Zeng\textsuperscript{1,2\ding{44}},} 
  \textbf{Zonghao Chen\textsuperscript{1,2\ding{44}},}
  \textbf{Ke Chen\textsuperscript{1,2},}
  \textbf{Lidan Shou\textsuperscript{1,2},}
  \textbf{Gang Chen\textsuperscript{1},}
  \textbf{Huan Li\textsuperscript{1,2\ding{41}}}\\
  \textsuperscript{1}The State Key Laboratory of Blockchain and Data Security, Zhejiang University \\
  \textsuperscript{2}Hangzhou High-Tech Zone (Binjiang) Institute of Blockchain and Data Security \\
  \small{\tt\{zj.cs, jiyicheng.cs, feiyangren, zbw.cs, 22521269, chenk, should, cg, lihuan.cs\}@zju.edu.cn} 
}

\begin{document}
\maketitle

\let\oldthefootnote\thefootnote
\renewcommand{\thefootnote}{}

\footnotemark
\footnotetext{\textsuperscript{\ding{44}}Equal contribution. \textsuperscript{\ding{41}}Corresponding author.}

\let\thefootnote\oldthefootnote
\setcounter{footnote}{0}


\begin{abstract}
Large Vision-Language Models (LVLMs) enable sophisticated reasoning over images and videos, yet their inference is hindered by a systemic efficiency barrier known as \emph{visual token dominance}. This overhead is driven by a multi-regime interplay between high-resolution feature extraction, quadratic attention scaling, and memory bandwidth constraints. We present a systematic taxonomy of efficiency techniques structured around the inference lifecycle, consisting of \emph{encoding}, \emph{prefilling}, and \emph{decoding}. Unlike prior reviews focused on isolated optimizations, we analyze the end-to-end pipeline to reveal how upstream decisions dictate downstream bottlenecks, covering compute-bound visual encoding, the intensive prefilling of massive contexts, and the ``visual memory wall'' in bandwidth-bound decoding. 
By decoupling the efficiency landscape into the axes of shaping information density, managing long-context attention, and overcoming memory limits, this work provides a structured analysis of how isolated optimizations compose to navigate the trade-off between visual fidelity and system efficiency. The survey concludes by outlining four future frontiers supported by pilot empirical insights, including hybrid compression based on functional unit sensitivity, modality-aware decoding with relaxed verification, progressive state management for streaming continuity, and stage-disaggregated serving through hardware-algorithm co-design.
Our literature repository is at {\href{https://github.com/SuDIS-ZJU/Efficient-LVLMs-Inference}{{\texttt{https://github.com/SuDIS-ZJU/Efficien\\t-LVLMs-Inference}}}}.
\end{abstract}

\section{Introduction}
\label{sec:intro}

Large Vision-Language Models (LVLMs)~\cite{wang2024qwen2,an2025llavaonevision15,wang2025internvl35} have evolved from research artifacts into the infrastructure for complex multimodal reasoning. 
However, as these models scale to process fine-grained visual inputs and long-form video streams, they encounter a systemic efficiency barrier: \emph{visual token dominance}~\cite{yang2024visionziplongerbetternecessary,tao2025dycoke,liu2025video}. 
Unlike text-only inputs, visual data yields orders of magnitude more tokens, pushing inference into a regime constrained not merely by compute cycles, but by the quadratic scaling of attention and the ``visual memory wall''~\footnote{For instance, a \texttt{Qwen2.5-VL-72B} processing 20 images already exceeds 40K tokens and 13 GB of cache, while a 5-second 720p video surpasses 50K tokens and 16 GB.}~\citep{wan2024look,li2025madakv,wang2025sparsemm}.

The central thesis of this survey is that LVLM inference is not a monolithic workload, but a dynamic pipeline traversing three distinct hardware regimes:
i) \textbf{\emph{Encoding}} (specifically \emph{visual} encoding) is compute-bound by high-resolution feature extraction; 
ii) \textbf{\emph{Prefilling}} suffers from the quadratic complexity of massive visual contexts; and 
iii) \textbf{\emph{Decoding}} hits the memory wall due to static, bandwidth-consuming Key-Value (KV) caches.
Optimizing one stage in isolation often shifts the bottleneck elsewhere without improving end-to-end latency.

Despite the surge in interest, the current literature remains fragmented.
Prior reviews have predominantly focused on isolated verticals, such as token compression techniques~\citep{shao2025tokenstalkmuchsurvey} or efficient architectures for specific modalities~\citep{ELLMSurvey,zhang-etal-2024-mm}~\footnote{\Cref{app:related_survey} provides a detailed related survey discussion.}. 
These works, however, overlook the systemic interconnectivity of the inference pipeline. They lack a holistic view of how upstream decisions (e.g., encoder resolution) dictate downstream bottlenecks (e.g., decoding bandwidth), leaving a gap in understanding end-to-end efficiency.

This survey bridges this gap by advancing a unified, \textbf{\emph{stage-wise taxonomy}} of efficient LVLM inference. 
We decouple the efficiency landscape into three critical axes: \emph{shaping information density} (encoding), \emph{managing long-context attention} (prefilling), and \emph{overcoming memory bandwidth limits} (decoding).
This framework provides a structured lens to evaluate how isolated optimizations compose, helping researchers navigate the trade-off between visual fidelity and system efficiency.

\begin{figure*}[t] 
  \centering
  \begin{minipage}{0.32\textwidth}
    \centering
    \includegraphics[width=\linewidth]{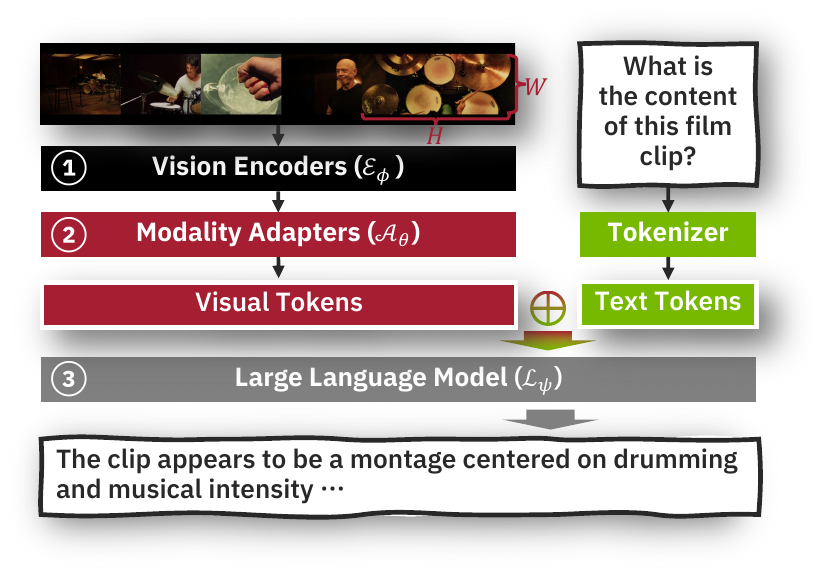}
    \caption{Three-stage pipeline for \\ LVLM inference.}
    \label{fig:arc}
  \end{minipage}
  \hspace{-0.34cm} 
  \begin{minipage}{0.69\textwidth}
    \centering
    \includegraphics[width=\linewidth]{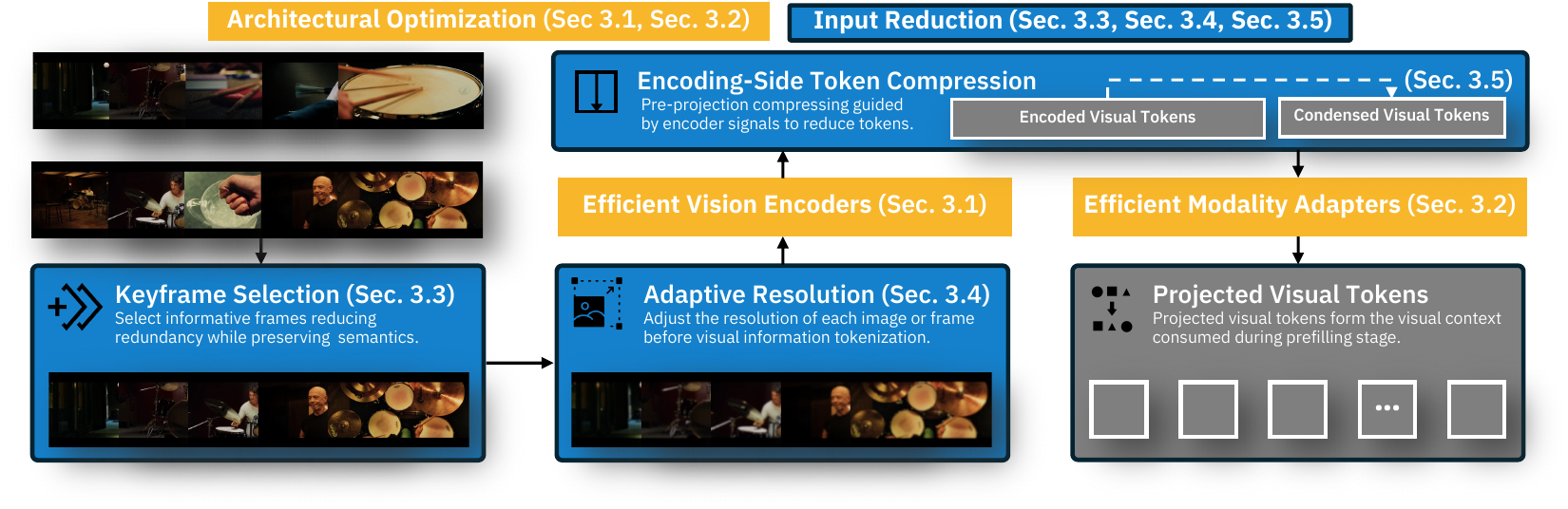}
    \centering
    \caption{Efficient encoding workflow: architectural optimization (\Cref{ssec:efficient_encoders}, \Cref{ssec:efficient_adapters}) and input reduction (\Cref{ssec:keyframe_selection}, \Cref{ssec:visual_downsampling}, \Cref{ssec:encoding_token_compression}).}
    \label{fig:encoding}
  \end{minipage}
\end{figure*}

\section{Preliminaries and Inference Dynamics}
\label{sec:problem_formulation}

LVLMs encounter unique efficiency bottlenecks compared to Large Language Models (LLMs), primarily due to the massive visual inputs. We formalize the canonical LVLM architecture (\Cref{ssec:canonical_arch}) and analyze its inference dynamics through a ``physics of computing'' lens (\Cref{ssec:physics_bottlenecks}), mapping hardware bottlenecks to user-centric metrics to structure the survey (\Cref{ssec:roadmap}).

\subsection{The Canonical LVLM Architecture}
\label{ssec:canonical_arch}

LVLMs typically adopt a \emph{three-stage pipeline} (\Cref{fig:arc}) that connects a vision encoder to an LLM.\footnote{Detailed component implementations and model taxonomy are provided in \Cref{app:architecture}.}
Given a multimodal tuple $(\mathbf{V}, \mathbf{T})$ comprising raw visual input\footnote{$F$ frames with $(H \cdot W)$ resolution and RGB channels.} $\mathbf{V} \in \mathbb{R}^{F \times H \times W \times 3}$ and a text prompt $\mathbf{T}$ of $N_t$ tokens, the pipeline is formalized as:
\begin{enumerate}[label=\ding{\numexpr171+\arabic*\relax}, itemsep=2pt, topsep=2pt, leftmargin=*]
    \item \textit{Visual Encoding.} 
    The encoder $\mathcal{E}_\phi$ (with parameters $\phi$) processes $\mathbf{V}$ into patch embeddings $\mathbf{X}_v \in \mathbb{R}^{N_p \times D_v}$ with $N_p$ the output patch number\footnote{For single-frame inputs ($F=1$), $N_p = (H \cdot W)/P^2$ with patch size $(P \times P)$; for videos with $F \geq 2$, $N_p$ depends on keyframe selection (\Cref{ssec:keyframe_selection}), adaptive resolution (\Cref{ssec:visual_downsampling}), and other compression strategies (e.g., frame pooling and Q-Former).} and $D_v$ the vision channel dimension. 

    \item \textit{Modality Alignment.} 
    A modality adapter $\mathcal{A}_\theta$ (with parameters $\theta$) maps $\mathbf{X}_v$ into the LLM's latent space, yielding visual context $\mathbf{H}_v = \mathcal{A}_\theta(\mathbf{X}_v) \in \mathbb{R}^{N_v \times D_{\mathcal{L}}}$ with $D_{\mathcal{L}}$ the LLM hidden dimension.
    The effective token count $N_v$ varies by projection strategy (e.g., pooling), defined by the compression ratio $r = N_v / N_p$.

    \item \textit{Autoregressive Generation.}
    The LLM backbone $\mathcal{L}_\psi$ (with parameters $\psi$) concatenates visual and text embeddings into a joint context $\mathbf{C} = [\mathbf{H}_v; \mathbf{H}_t]$ (where $\mathbf{H}_t \in \mathbb{R}^{N_t \times D_{\mathcal{L}}}$ represents the prompt of $N_t$ textual tokens) to generate the output response $\mathbf{Y} = (y_1, \dots, y_{N_{o}})$ of length $N_{o}$ autoregressively:
    \begin{equation}
        p(\mathbf{Y} \mid \mathbf{C}) = \prod\nolimits_{k=1}^{N_{o}} p(y_k \mid \mathbf{C}, y_{<k}; \psi).
    \end{equation}
\end{enumerate}

Here, a defining characteristic is the \textbf{\emph{visual token dominance}}:
the visual content ($N_v \approx$ 576 -- 4,000+) significantly exceeds standard text prompts ($N_v \gg N_t$). 
This structural imbalance dictates the inference bottlenecks analyzed below.%

\subsection{The Physics of Inference Bottlenecks}
\label{ssec:physics_bottlenecks}

We model the end-to-end inference latency as:
\begin{equation}\label{equation:total_latency}
    \tau_{\text{total}} = \tau_{\text{ENC}} + \tau_{\text{PFL}} + N_{o} \cdot \tau_{\text{DEC}},
\end{equation}
where the first two terms contribute to Time-to-First-Token (TTFT) at encoding and prefilling, respectively, and $\tau_{\text{DEC}}$ determines Time-Per-Output-Token (TPOT) at decoding.
To identify bottlenecks, we apply the \emph{Roofline model}~\footnote{A detailed Roofline analysis is provided in~\Cref{app:roofline_details}.}, which bounds performance based on the workload's \emph{arithmetic intensity} $\mathcal{I}$ (FLOPs$/$Byte).
A stage is \textbf{\emph{compute-bound}} if $\mathcal{I} \geq \pi_{\text{peak}}/\beta_{\text{mem}}$, saturating the peak compute throughput $\pi_{\text{peak}}$; otherwise, it is \textbf{\emph{memory-bound}}, throttled by the memory bandwidth $\beta_{\text{mem}}$.
%
As summarized in~\Cref{tab:bottlenecks}, encoding is compute-bound with high arithmetic intensity from dense matrix operations; prefilling exhibits mixed behavior where both computation (quadratic attention) and memory I/O (KV cache materialization) can dominate; and decoding is strictly memory-bound due to the low arithmetic intensity of autoregressive token generation. Understanding these stage-specific bottlenecks is crucial for targeting optimization efforts.

\begin{table*}[!t]
    \centering
    \caption{Hardware bottlenecks, arithmetic intensity, and complexity dynamics across the three inference stages.}
    \label{tab:bottlenecks}
    \resizebox{\textwidth}{!}{%
        \begin{tabular}{l|ccc}
        \toprule
        \textbf{Characteristic} & 
        \textbf{Encoding Stage} & 
        \textbf{Prefilling Stage} & 
        \textbf{Decoding Stage} \\
        \midrule
        \textbf{Primary Metric} & TTFT & TTFT & TPOT \\
        \rowcolor{hanred!40}
        \textbf{Bottleneck} & \textbf{Compute-Bound} & \textbf{Compute \& Memory Bound} & \textbf{Memory-Bound} \\
        \textbf{Arithmetic Intensity} & High ($\gg 1$) & Medium & Low ($\ll 1$) \\
        \rowcolor{gray!35}
        \textbf{Complexity (FLOPs)} & $\mathcal{O}(N_p \cdot D_v^2)$ & $\mathcal{O}((N_v+N_t)^2 \cdot D_{\mathcal{L}})$ & $\mathcal{O}((N_v+N_t) \cdot D_{\mathcal{L}})$ \\
        \textbf{LVLM Challenge} &
            High-res inputs ($N_p \uparrow$) surge FLOPs &
            ($N_v \gg N_t$) causes quadratic spikes &
            Static visual KV cache saturates VRAM \\
        \bottomrule
        \end{tabular}%
    }
\end{table*}

\paragraph{Encoding Stage: \emph{Compute-Bound}.}
This stage executes dense matrix multiplications over $N_p$ patches, a high-intensity workload (see \Cref{tab:bottlenecks}) strictly bounded by compute throughput:
\begin{equation}
    \tau_{\text{ENC}} \approx {\text{FLOPs}_{\text{ENC}}} / {\pi_{\text{peak}}}.
\end{equation}
The encoder produces $N_p$ patch embeddings, which are then projected by $\mathcal{A}_\theta$ to yield $N_v$ visual tokens entering the LLM. 
While encoding cost is constant per request (independent of $N_t$ or $N_{o}$), reducing $N_v$ yields \emph{cascading benefits}: it lowers prefilling complexity from $\mathcal{O}((N_v+N_t)^2)$ to $\mathcal{O}((N_v'+N_t)^2)$ where $N_v' < N_v$ (see \Cref{tab:bottlenecks}), and shrinks KV cache size linearly (\Cref{eq:kv_cache_size}).


\paragraph{Prefilling Stage: \emph{Compute \& Memory Bound}.}
This stage processes the context $\mathbf{C}$ to populate the initial Key-Value (KV) cache.
While attention computation is quadratic, the \textit{materialization} of the KV cache for massive visual tokens creates a heavy memory write burden. 
The latency is determined by the bottleneck resource:
\begin{equation}
    \tau_{\text{PFL}} \approx \max \left( \frac{\text{FLOPs}_{\text{attn}}}{\pi_{\text{peak}}},
                                           \frac{|\mathcal{K}\mathcal{V}|_{\text{PFL}}}{\beta_{\text{mem}}} \right),
\end{equation}
where $|\mathcal{K}\mathcal{V}|_{\text{PFL}} \approx 2 \cdot L \cdot (N_v + N_t) \cdot D_{\mathcal{L}}
\cdot r_{\text{kv}} \cdot \mathcal{P}$ represents the bytes written to HBM. Here, $L$ is the number of layers,
$\mathcal{P}$ is the element size in bytes, and $r_{\text{kv}}$ denotes the ratio of KV heads to Query heads
(i.e., $r_{\text{kv}}=1$ for MHA, $r_{\text{kv}} < 1$ for GQA/MQA). Unlike text-only prefilling, a large $N_v$ can push this
stage towards the memory wall.


\paragraph{Decoding Stage: \emph{Memory-Bound}.}
Generating each output token necessitates streaming the model weights $\psi$ and the accumulated KV cache from HBM to on-chip SRAM.
The KV cache size at generation step $i$ ($1 \le i \le N_{o}$) is dynamic:
\begin{equation}
\label{eq:kv_cache_size}
    |\mathcal{K}\mathcal{V}|_{i} \approx 2 \cdot L \cdot (N_v + N_t + i) \cdot D_{\mathcal{L}}
                                            \cdot r_{\text{kv}} \cdot \mathcal{P}.
\end{equation}
This stage is strictly memory-bound due to low arithmetic intensity (batch size $\approx 1$),
with the single-step latency $\tau_{\text{DEC}}^{(i)}$ and total decoding latency $\tau_{\text{DEC}}$
defined as~\footnote{We assume sufficient single-GPU memory capacity.
Thus, Tensor Parallelism across $N_{\text{gpu}}$ devices linearly scales the effective bandwidth to
$N_{\text{gpu}} \cdot \beta_{\text{mem}}$. However, since the arithmetic intensity remains unchanged,
the decoding process persists as strictly memory-bound on each individual device.}:
\begin{equation}
\begin{split}
    \tau_{\text{DEC}} &= \sum\nolimits_{i=1}^{N_{o}} \tau_{\text{DEC}}^{(i)}, \\
    \text{where } \quad \tau_{\text{DEC}}^{(i)} &\approx \big({|\psi| + |\mathcal{K}\mathcal{V}|_{i}} \big) / {\beta_{\text{mem}}}.
\end{split}
\end{equation}
Here, $|\psi|$ is the model weights size.
The \textit{visual memory wall} arises because the visual component $|\mathcal{K}\mathcal{V}|_v$
(where $|\mathcal{K}\mathcal{V}|_v \propto N_v \cdot L \cdot D_{\mathcal{L}}$) necessitates the repeated loading of massive static states, dominating memory bandwidth consumption throughout the entire
generation process ($N_{o}$ generation steps).

\definecolor{hidden-red}{RGB}{241, 120, 139}
\definecolor{hidden-blue}{RGB}{194,232,245}
\definecolor{hidden-orange}{RGB}{243,202,120}
\definecolor{hidden-green}{RGB}{34,139,34}
\definecolor{hidden-purple}{RGB}{204,176,252}
\definecolor{hidden-pink}{RGB}{255,245,247}
\definecolor{hidden-black}{RGB}{20,68,106}
\definecolor{purple}{RGB}{144,153,196}
\definecolor{yellow}{RGB}{255,228,123}
\definecolor{hidden-yellow}{RGB}{255,248,203}
\definecolor{tkcolor}{RGB}{224,223,255}
\definecolor{darkblue}{rgb}{0, 0.40, 0.75}
\definecolor{gray}{RGB}{231,231,231}
\definecolor{yellow-han}{RGB}{248,183,43}
\definecolor{green-han}{RGB}{118,185,0}
\definecolor{blue-han}{RGB}{21,129,204}
\definecolor{gray-han}{RGB}{128,128,128}
\definecolor{red-han}{RGB}{165,30,50}

\tikzstyle{my-box}=[
rectangle,
draw=hidden-black,
rounded corners,
text opacity=1,
minimum height=1.5em,
minimum width=5em,
inner sep=2pt,
align=center,
fill opacity=.5,
]
\tikzstyle{leaf3}=[
my-box,
minimum height=1.5em,
fill=yellow!32,
text=black,
align=left,
font=\normalsize,
inner xsep=5pt,
inner ysep=4pt,
align=left,
text width=45em,
]
\tikzstyle{leaf6}=[
my-box,
minimum height=1.5em,
fill=hidden-purple,
draw=hidden-purple,
text=black,
align=left,
font=\normalsize,
inner xsep=5pt,
inner ysep=4pt,
]
\tikzstyle{leaf4}=[
my-box,
minimum height=1.5em,
fill=hidden-blue,
draw=hidden-blue,
text=black,
align=left,
font=\normalsize,
inner xsep=5pt,
inner ysep=4pt,
]
\tikzstyle{leaf2}=[
my-box,
minimum height=1.5em,
fill=hidden-green!20,
text=black,
align=left,
font=\normalsize,
inner xsep=5pt,
inner ysep=4pt,
]
\tikzstyle{leaf}=[
my-box,
minimum height=1.5em,
fill=hidden-red,
draw=hidden-red,
text=black,
align=left,
font=\normalsize,
inner xsep=5pt,
inner ysep=4pt,
]
\tikzstyle{leaf5}=[
my-box,
minimum height=1.5em,
fill=darkblue!15,
text=black,
align=left,
font=\normalsize,
inner xsep=5pt,
inner ysep=4pt,
]

\tikzstyle{leaf1_han}=[
my-box,
minimum height=1.5em,
fill=green-han!35,
draw=green-han!35,
text=black,
align=left,
font=\normalsize,
inner xsep=5pt,
inner ysep=4pt,
]

\tikzstyle{leaf2_han}=[
my-box,
minimum height=1.5em,
fill=blue-han!35,
draw=blue-han!35,
text=black,
align=left,
font=\normalsize,
inner xsep=5pt,
inner ysep=4pt,
]

\tikzstyle{leaf3_han}=[
my-box,
minimum height=1.5em,
fill=yellow-han!35,
draw=yellow-han!35,
text=black,
align=left,
font=\normalsize,
inner xsep=5pt,
inner ysep=4pt,
]

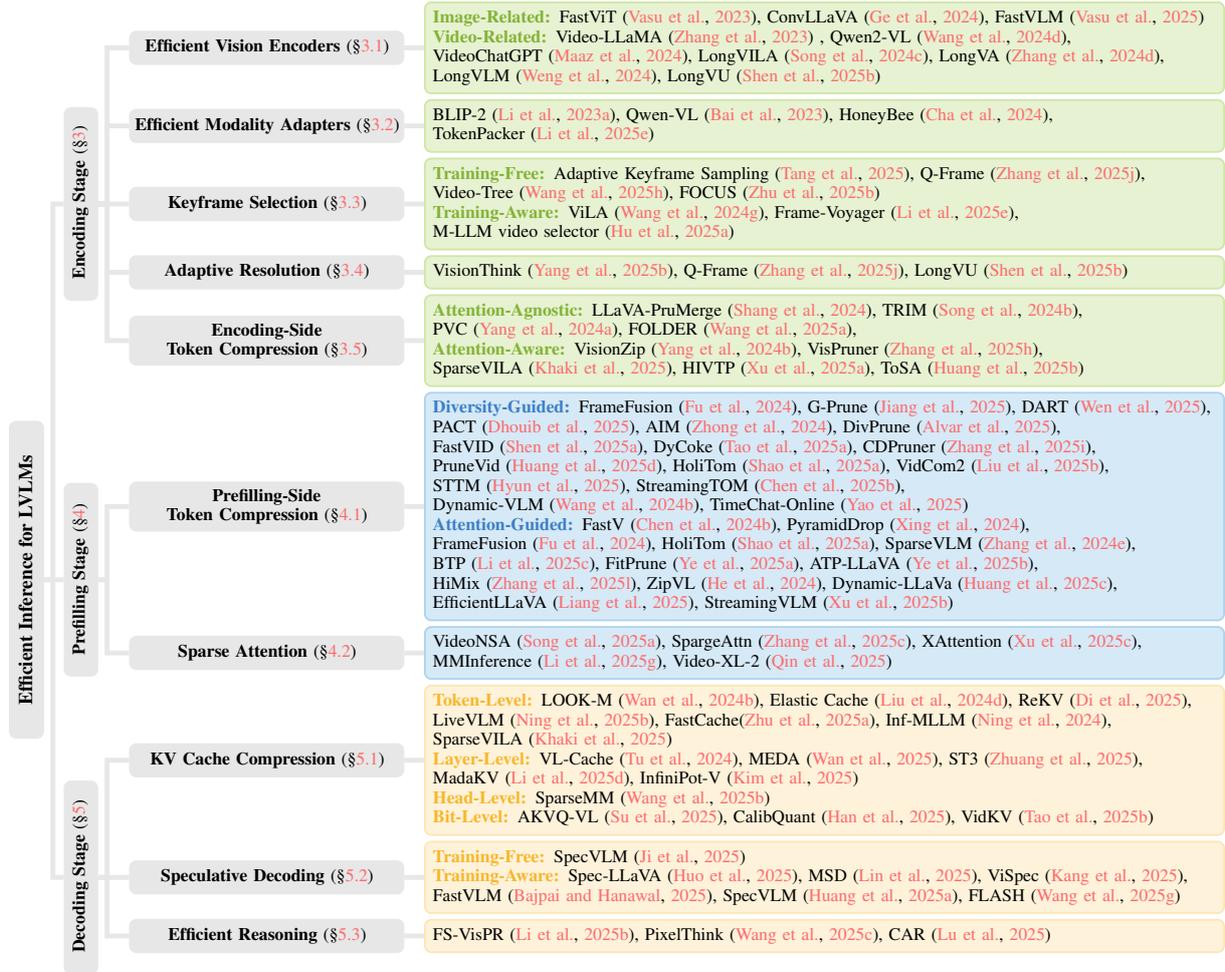
\begin{figure*}[!t]
\vspace{-2mm}
\centering
\resizebox{1\textwidth}{!}{%
    \begin{forest}
        forked edges,
        for tree={
            grow=east,
            reversed=true,
            anchor=base west,
            parent anchor=east,
            child anchor=west,
            base=left, 
            font=\large,
            rectangle,
            draw=gray,
            fill=gray,
            rounded corners,
            align=left, 
            minimum width=4em,
            edge+={gray, line width=3pt},
            s sep=3pt,
            inner xsep=2pt,
            inner ysep=4pt,
            line width=1.1pt,
            ver/.style={rotate=90, child anchor=north, parent anchor=south, anchor=center},
        },
        where level=1{text width=10.5em,font=\normalsize,}{},
        where level=2{text width=15em,font=\normalsize,}{},
        where level=3{text width=20em,font=\normalsize,}{},
        where level=4{text width=50em,font=\normalsize,}{},
        [\parbox{16em}{\centering \textbf{Efficient Inference for LVLMs}}, ver
            [\quad \textbf{Encoding Stage}~(\S\ref{sec:encoding})\quad, ver,
                [\parbox{15em}{\centering \textbf{Efficient Vision Encoders}~(\S\ref{ssec:efficient_encoders})}, anchor=west
                    [
                    \textbf{\textcolor{hangreen}{Image-Related:}} FastViT~\cite{vasu2023fastvit}{,} ConvLLaVA~\cite{ge2024convllava}{,} FastVLM~\cite{vasu2025fastvlm} \\
                    \textbf{\textcolor{hangreen}{Video-Related:}}
                    Video-LLaMA~\cite{zhang2023video}
                    {,}
                    Qwen2-VL~\cite{wang2024qwen2}{,}\\ VideoChatGPT~\cite{maaz2024video}{,} LongVILA~\cite{song2024moviechat}{,} LongVA~\cite{zhang2024long}{,}\\ LongVLM~\cite{weng2024longvlm}{,}
                    LongVU~\cite{shenlongvu}, leaf1_han, text width=44em, anchor=west
                    ]
                ]
                [\parbox{15em}{\centering \textbf{Efficient Modality Adapters}~(\S\ref{ssec:efficient_adapters})}, anchor=west
                    [
                    BLIP-2~\cite{li2023blip}{,} 
                    Qwen-VL~\cite{bai2023qwen}{,}
                    HoneyBee~\cite{cha2024honeybee}{,}\\
                    TokenPacker~\cite{li2025tokenpacker},
                    leaf1_han, text width=44em, anchor=west
                    ]
                ]
                [\parbox{15em}{\centering \textbf{Keyframe Selection}~(\S\ref{ssec:keyframe_selection})}, anchor=west, calign with current
                    [
                    \textbf{\textcolor{hangreen}{Training-Free:}} Adaptive Keyframe Sampling~\cite{tang2025adaptive}{,}
                    Q-Frame~\cite{zhang2025q}{,} \\ Video-Tree~\cite{wang2025videotree}{,} FOCUS~\cite{zhu2025focus}\\
                    \textbf{\textcolor{hangreen}{Training-Aware:}} ViLA~\cite{wang2024vila}{,} Frame-Voyager~\cite{li2025tokenpacker}{,}\\ M-LLM video selector~\cite{hu2025m},
                    leaf1_han, text width=44em, anchor=west
                    ]
                ]
                [\parbox{15em}{\centering \textbf{Adaptive Resolution}~(\S\ref{ssec:visual_downsampling})}, anchor=west
                [
                VisionThink~\cite{yang2025visionthink}{,}  
                    Q-Frame~\cite{zhang2025q}{,} LongVU~\cite{shenlongvu},
                    leaf1_han, text width=44em, anchor=west
                ]
                ]
                [\parbox{15em}{\centering \textbf{Encoding-Side\\Token Compression}~(\S\ref{ssec:encoding_token_compression})}, anchor=west
                    [
                    \textbf{\textcolor{hangreen}{Attention-Agnostic:}} LLaVA-PruMerge~\cite{shang2024llavaprumerge}{,} 
                    TRIM~\cite{song2024moresimpleeffectivetoken}{,} 
                    \\PVC~\cite{yang2024pvcprogressivevisualtoken}{,}  FOLDER~\cite{wang2025folderacceleratingmultimodallarge}{,} \\
                    \textbf{\textcolor{hangreen}{Attention-Aware:}} VisionZip~\cite{yang2024visionziplongerbetternecessary}{,} VisPruner~\cite{zhang2025textvisualattentionexploitingvisual}{,}\\ SparseVILA~\cite{khaki2025sparsevila}{,} HIVTP~\cite{xu2025hivtptrainingfreemethodimprove}{,} ToSA~\cite{huang2025tosatokenmergingspatial},
                    leaf1_han, text width=44em, anchor=west
                    ]
                ]
            ]
            [\quad \textbf{Prefilling Stage}~(\S\ref{sec:prefilling})\quad, ver, calign with current
                [\parbox{15em}{\centering \textbf{Prefilling-Side\\Token Compression}~(\S\ref{ssec:prefilling_token_compression})}, anchor=west
                    [
                    \textbf{\textcolor{hanblue}{Diversity-Guided:}} 
                    FrameFusion~\cite{fu2024framefusion}{,}
                    G-Prune~\cite{jiang2025kind}{,}
                    DART~\cite{wen2025stop}{,} \\
                    PACT~\cite{dhouib2025pact}{,}
                    AIM~\cite{zhong2025aim}{,} 
                    DivPrune~\cite{alvar2025divprune}{,} 
                     \\
                    FastVID~\cite{shen2025fastvid}{,}
                    DyCoke~\cite{tao2025dycoke}{,} 
                    CDPruner~\cite{zhang2025beyond}{,} 
                    \\
                    PruneVid~\cite{huang2025prunevid}{,}
                    HoliTom~\cite{shao2025holitom}{,} 
                    VidCom2~\cite{liu2025video}{,} \\
                    STTM~\cite{hyun2025multi}{,} 
                    StreamingTOM~\cite{chen2025streamingtom}{,} \\
                    Dynamic-VLM~\cite{wang2025dynamic}{,} 
                    TimeChat-Online~\cite{yao2025timechat} \\
                    \textbf{\textcolor{hanblue}{Attention-Guided:}} FastV~\cite{chen2024image}{,} PyramidDrop~\cite{xing2024pyramiddrop}{,} \\
                    FrameFusion~\cite{fu2024framefusion}{,}
                    HoliTom~\cite{shao2025holitom}{,}
                    SparseVLM~\cite{zhang2024sparsevlm}{,} \\ 
                    BTP~\cite{li2025balanced}{,} FitPrune~\cite{ye2025fit}{,} ATP-LLaVA~\cite{ye2025atp}{,} \\ 
                    HiMix~\cite{zhang2025himix}{,} ZipVL~\cite{he2024zipvl}{,}
                    Dynamic-LLaVa~\cite{huang2025dynamic}{,} \\
                    EfficientLLaVA~\cite{liang2025efficient}{,} StreamingVLM~\cite{xu2025streamingvlm},
                    leaf2_han, text width=44em, anchor=west
                    ]
                ]
                [\parbox{15em}{\centering \textbf{Sparse Attention}~(\S\ref{ssec:sparse_attention})}, anchor=west
                    [
                    VideoNSA~\cite{song2025videonsa}{,} 
                    SpargeAttn~\cite{zhang2025spargeattn}{,} 
                    XAttention~\cite{xu2025xattention}{,} \\MMInference~\cite{li2025mminference}{,}
                    Video-XL-2~\cite{qin2025video},
                    leaf2_han, text width=44em, anchor=west
                    ]
                ]
            ]
            [\quad \textbf{Decoding Stage}~(\S\ref{sec:decoding})\quad, ver,
                [\parbox{15em}{\centering \textbf{KV Cache Compression}~(\S\ref{ssec:kv_compression})}, anchor=west
                    [
                    \textbf{\textcolor{hanyellow}{Token-Level:}} LOOK-M~\cite{wan2024look}{,} Elastic Cache~\cite{liu2024efficient}{,} 
                    ReKV~\cite{di2025streaming}{,} 
                    \\ LiveVLM~\cite{ning2025livevlm}{,} 
                    FastCache\cite{zhu2025fastcache}{,} Inf-MLLM~\cite{ning2024inf}{,}
                    \\ 
                    SparseVILA~\cite{khaki2025sparsevila} 
                    \\
                    \textbf{\textcolor{hanyellow}{Layer-Level:}} VL-Cache~\cite{tu2024vl}{,} MEDA~\cite{wan2025meda}{,} ST3~\cite{zhuang2025st3}{,} \\ MadaKV~\cite{li2025madakv}{,} InfiniPot-V~\cite{kim2025infinipot} \\ 
                    \textbf{\textcolor{hanyellow}{Head-Level:}} SparseMM~\cite{wang2025sparsemm}\\
                    \textbf{\textcolor{hanyellow}{Bit-Level:}} AKVQ-VL~\cite{su2025akvq}{,} CalibQuant~\cite{han2025calibquant}{,} VidKV~\cite{tao2025plug},
                    leaf3_han, text width=44em, anchor=west
                    ]
                ]
                [\parbox{15em}{\centering \textbf{Speculative Decoding}~(\S\ref{ssec:sepculative_decoding})}, anchor=west,calign with current
                    [
                    \textbf{\textcolor{hanyellow}{Training-Free:}} SpecVLM~\cite{ji2025specvlm} \\
                    \textbf{\textcolor{hanyellow}{Training-Aware:}} Spec-LLaVA~\cite{huo2025spec}{,} MSD~\cite{lin2025speculative}{,} ViSpec~\cite{kang2025vispec}{,} \\ FastVLM~\cite{bajpai2025fastvlm}{,} SpecVLM~\cite{huang2025specvlm}{,} FLASH~\cite{wang2025flash},
                    leaf3_han, text width=44em, anchor=west
                    ]
                ]
                [\parbox{15em}{\centering \textbf{Efficient Reasoning}~(\S\ref{ssec:efficient_reasoning})}, anchor=west
                    [
                    FS-VisPR~\cite{li2025adaptive}{,} PixelThink~\cite{wang2025pixelthink}{,} CAR~\cite{lu2025prolonged},
                    leaf3_han, text width=44em, anchor=west
                    ]
                ]
            ]
        ]
    \end{forest}%
}
\caption{A stage-aware taxonomy of efficient LVLM inference. 
We categorize techniques by their intervention stage and optimization mechanism. This framework maps research to their target hardware bottlenecks, elucidating {WHERE} in the lifecycle and {HOW} via specific algorithms computational redundancy is reduced.}
\label{fig:taxonomy}
\vspace{-1.0em}
\end{figure*}

\subsection{Survey Organization}
\label{ssec:roadmap}

Given this bottleneck analysis, we organize the remainder of the survey around the stage-aware taxonomy illustrated in \Cref{fig:taxonomy}:
\Cref{sec:encoding} examines upstream techniques on architectural optimization and input reduction to minimize $\tau_{\text{ENC}}$ and reduce $N_v$; \Cref{sec:prefilling} focuses on mitigating quadratic computation via token compression and sparse attention; and \Cref{sec:decoding} addresses the memory-bound decoding stage via KV cache optimization, speculative execution, and efficient reasoning. 
Crucially, we distill empirical insights from each section into a set of \textbf{Key Takeaways} in \Cref{app:takeaway}, which serve as the foundation for future directions discussed in \Cref{sec:futurework}. As essential supplementary references, \Cref{app:architecture} details architectural taxonomies while \Cref{app:sys} presents system-level serving and evaluation frameworks.

\section{Efficiency Techniques at Encoding}
\label{sec:encoding}

%
Guided by the workflow in \Cref{fig:encoding}, this section surveys efficiency techniques for the LVLM encoding stage, structured into two strategic axes:
i) \emph{architectural optimization} focuses on designing vision encoders $\mathcal{E}_\phi$  (\Cref{ssec:efficient_encoders}) and adapters $\mathcal{A}_\theta$ (\Cref{ssec:efficient_adapters}) to minimize the on-model tokenization latency $\tau_{\text{ENC}}$; and
ii) \emph{input reduction} explores optimized visual token representations to reduce the number of visual tokens $N_v$ entering the downstream pipeline, including keyframe selection (\Cref{ssec:keyframe_selection}), adaptive resolution  (\Cref{ssec:visual_downsampling}), and encoding-side token compression  (\Cref{ssec:encoding_token_compression}).


\subsection{Efficient Vision Encoders}
\label{ssec:efficient_encoders}

The vision encoder $\mathcal{E}_\phi$ acts as the upstream efficiency regulator, governing the initial visual token density $N_v$ that propagates through the pipeline. 

\paragraph{Image-Related.}
Recent architectures optimize backbone efficiency through structural reparameterization (FastViT~\cite{vasu2023fastvit}) and distillation (EfficientViT-SAM~\cite{zhang2024accelerated}). To mitigate token bloat from high-resolution inputs, ConvLLaVA~\cite{ge2024convllava} and FastVLM~\cite{vasu2025fastvlm} employ hierarchical compression and hybrid encoding to generate compact feature sets.

\paragraph{Video-Related.}
Approaches here focus on temporal adaptation and scalability.
Video-LLaMA~\cite{zhang2023video} propose a video Q-Former to assemble a pre-trained image encoder into video encoder, while Qwen2-VL~\cite{wang2024qwen2} implements Native Dynamic Resolution for adaptive token generation.
VideoChatGPT~\cite{maaz2024video} enhances image encoders to capture spatiotemporal representations in videos.
For long-context scenarios, MovieChat~\cite{song2024moviechat},  LongVA~\cite{zhang2024long}, LongVLM~\cite{weng2024longvlm},
LongVILA~\cite{chenlongvila}, and LongVU~\cite{shenlongvu} leverage context extension and supervised fine-tuning to support extended temporal encoding.




\MySummaryBox{The field is pivoting from static, fixed-resolution backbones toward dynamic, density-aware architectures.
Future architectures will likely favor end-to-end learnable compressors to maximize this upstream efficiency.}

\subsection{Efficient Modality Adapters}
\label{ssec:efficient_adapters}
The modality adapter $\mathcal{A}_\theta$ semantically aligns the vision encoder's outputs with the LLM backbone.
Baseline architectures like LLaVA~\cite{liu2023llava,liu2023improvedllava} employ simple MLPs. While computationally inexpensive, this one-to-one mapping prevents token reduction, causing visual token count $N_v$ to scale linearly with input resolution.
To tackle the token explosion problem, BLIP-2~\cite{li2023blip} bridges the modality gap with a lightweight Q-Former. Recent works introduce resampler~\cite{bai2023qwen} or abstractor~\cite{cha2024honeybee} to enforce compactness. TokenPacker~\cite{li2025tokenpacker} further refines this via a coarse-to-fine injection scheme, "packing" enriched visual semantics into fewer tokens.



\MySummaryBox{The adapter is evolving from a passive bridge to an active information bottleneck, prioritizing high-density latent representations over raw feature preservation to mitigate downstream computational load.}

\subsection{Keyframe Selection}
\label{ssec:keyframe_selection}

Keyframe selection acts as a pre-encoding filter, discarding redundant frames from $\mathbf{V}$ to minimize the computational load on the vision encoder $\mathcal{E}_\phi$. We categorize these strategies by their optimization substrate: training-free with heuristic metrics versus training-aware with learnable policies.

\paragraph{Training-Free Selection.}
This paradigm decouples selection from model training, deploying frozen encoders as plug-and-play scorers to rank frames by semantic relevance~\cite{yu2023self,ranasinghe2024understanding,liang2024keyvideollm}.
Beyond thresholding, recent works introduce structural priors:
Adaptive keyframe sampling~\cite{tang2025adaptive} jointly optimizes prompt relevance and temporal coverage via a split-and-judge policy.
Q-Frame~\cite{zhang2025q} employs the Gumbel-Max trick based on a text-image matching network for efficient probabilistic sampling;
VideoTree~\cite{wang2025videotree} constructs a hierarchical tree to extract query-relevant details from long videos in a coarse-to-fine manner;
and FOCUS~\cite{zhu2025focus} formulates selection as a combinatorial pure-exploration problem in multi-armed bandits.

\paragraph{Training-Aware Selection.}
This paradigm, conversely, treats selection as a learnable policy optimized end-to-end for downstream performance.
ViLA~\cite{wang2024vila} learns a text-guided ``Frame-Prompter'' to identify question-related frames that maximize video QA accuracy, while Frame-Voyager~\cite{yu2024frame} minimizes the combination loss against ground-truth answers.
Others, like the M-LLM video selector~\cite{hu2025m}, employ explicit cross-entropy-based supervision from spatial and temporal signals.



\MySummaryBox{The trade-off is pragmatic: heuristics offer portability for open-domain deployment, whereas learnable policies can combat reasoning biases and select frames conditioned on the query.}


\subsection{Adaptive Resolution}
\label{ssec:visual_downsampling}

Adaptive resolution optimizes the upstream information budget by modulating input fidelity prior to tokenization.
For static visual inputs, methods like VisionThink~\cite{yang2025visionthink} and ViCO~\cite{cui2025vicotrainingstrategysemantic} implement complexity-aware scaling, dynamically adjusting resolution or selecting image compression ratios via multi-branch MLP connectors based on semantic difficulty of samples.
This logic extends to query-conditional resolution for videos: Q-Frame~\cite{zhang2025q} and LongVU~\cite{shenlongvu} maintain high-fidelity features strictly for query-relevant frames, while aggressively reducing background context via spatial pooling or downsampling.




\MySummaryBox{This marks a shift toward content-adaptive encoding, where the computational budget is dynamically allocated to high-value signals rather than wasted on uniform processing.}

\subsection{Encoding-Side Token Compression}
\label{ssec:encoding_token_compression}

This category reduces the visual token count $N_v$ immediately after encoding, operating independently of the LLM backbone. Techniques are categorized by their reliance on the encoder's internal signals.

\paragraph{Attention-Agnostic Compression.} 
These methods exploit the inherent spatial redundancy of visual patches using lightweight similarity metrics.
LLaVA-PruMerge~\cite{shang2024llavaprumerge} and TRIM~\cite{song2024moresimpleeffectivetoken} prune encoder-output tokens based on their similarities to the global [CLS] token or CLIP-based metrics.
PVC~\cite{yang2024pvcprogressivevisualtoken} adopts a progressive strategy by treating static images as pseudo-temporal sequences to filter redundant features. FOLDER~\cite{wang2025folderacceleratingmultimodallarge} integrates a plug-and-play merging module directly into the final blocks of the encoder.

\paragraph{Attention-Aware Compression.} 
These methods~\cite{han2026filter} utilize the encoder's self-attention maps as proxies for feature saliency.
VisionZip~\cite{yang2024visionziplongerbetternecessary}, VisPruner~\cite{zhang2025textvisualattentionexploitingvisual}, and SparseVILA~\cite{khaki2025sparsevila} directly derive importance scores from attention matrices to retain high-value tokens. Extensions like HIVTP~\cite{xu2025hivtptrainingfreemethodimprove} extract attention maps from intermediate layers of the vision encoder for early filtering, while ToSA~\cite{huang2025tosatokenmergingspatial} combines semantic attention with spatial proximity to perform spatially-aware token merging.



\MySummaryBox{Unlike prefilling-stage compression (\Cref{sec:prefilling}), these techniques are upstream and prompt-agnostic, compressing visual data solely based on intrinsic visual properties without interaction from the textual query or LLM weights.}




\section{Efficiency Techniques at Prefilling}
\label{sec:prefilling}

This section surveys efficiency techniques for LVLM prefilling, where causal self-attention processes massive visual contexts to materialize the KV cache.
As the primary determinant of TTFT (see \Cref{equation:total_latency}), prefilling latency $\tau_{\text{PFL}}$ acts as a hard gate on responsiveness.
To mitigate this bottleneck, we structure the landscape into two strategic axes:
i) \emph{prefilling-side token compression} (\Cref{ssec:prefilling_token_compression}) that aims to reduce the quantity of visual tokens ($N_v$) during prefilling, and ii) \emph{sparse attention} (\Cref{ssec:sparse_attention}) that reduces computational complexity of the attention mechanism itself.

\begin{table*}[t]
\centering
\tiny
\renewcommand{\arraystretch}{1.1}
\begin{tabular}{@{}llccl@{}}
\toprule
\textbf{Category} & \textbf{Method} & \textbf{Input Modality} & \textbf{Training-Free} & \textbf{Key Strategy \& Insight} \\ \midrule
\multirow{16}{*}{\textbf{Diversity-Guided}} 
 & G-Prune~\cite{jiang2025kind} & General & Yes & Similarity graph \& information flow to retain representative tokens \\
 & PACT~\cite{dhouib2025pact} & General & Yes & Distance-bounded clustering \& merging redundant tokens \\
 & DivPrune~\cite{alvar2025divprune} & General & Yes & Max-Min diversity optimization for token subset selection \\
 & CDPruner~\cite{zhang2025beyond} & General & Yes & Determinantal Point Processes (DPP) \& conditional diversity \\
 & DART~\cite{wen2025stop} & General & Yes & Pivot-based duplication pruning \\ \cmidrule(l){2-5} 
 & DyCoke~\cite{tao2025dycoke} & Video & Yes & Plug-and-play module for temporal token merging \\
 & PruneVid~\cite{huang2025prunevid} & Video & Yes & Spatiotemporal token merging \\
 & AIM~\cite{zhong2025aim} & Video & Yes & Spatiotemporal token merging \\
 & FrameFusion~\cite{fu2024framefusion} & Video & Yes & Merges shallow-layer tokens based on adjacent frame similarity \\
 & FastVID~\cite{shen2025fastvid} & Video & Yes & Temporal segmentation \& density spatiotemporal pruning \\
 & HoliTom~\cite{shao2025holitom} & Video & Yes & Global redundancy-aware segmentation \& spatiotemporal merging \\
 & VidCom2~\cite{liu2025video} & Video & Yes & Dynamic compression based on frame uniqueness \\
 & STTM~\cite{hyun2025multi} & Video & Yes & Quadtree spatial transformation \& directed pairwise merging \\
 & Dynamic-VLM~\cite{wang2025dynamic} & Video & No & Dynamic compression architecture adapting to video length \\
 & StreamingTOM~\cite{chen2025streamingtom} & Streaming Video & Yes & Causal temporal reduction with fixed per-frame budget \\
 & TimeChat-Online~\cite{yao2025timechat} & Streaming Video & No & Differential token drop for redundant content filtering \\ \midrule
\multirow{10}{*}{\textbf{Attention-Guided}} 
 & FastV~\cite{chen2024image} & General & Yes & Learns attention patterns in early layers to prune in deep layers \\
 & PyramidDrop~\cite{xing2024pyramiddrop} & General & No & Multi-stage pruning using attention score ranking \\
 & HiMix~\cite{zhang2025himix} & General & No & Hierarchical vision injection via mixture attention \\
 & ZipVL~\cite{he2024zipvl} & General & Yes & Dynamic token sparsification based on attention scores \\
 & Dynamic-LLaVA~\cite{huang2025dynamic} & General & No & Dynamic vision-language context sparsification \\
 & EfficientLLaVA~\cite{liang2025efficient} & General & No & Few-shot pruning policy search via structural risk minimization \\
 & BTP~\cite{li2025balanced} & General & Yes & Multi-stage pruning with diversity and attention ranking \\
 & SparseVLM~\cite{zhang2024sparsevlm} & General & Yes & Pruning based on text-visual attention scores \\
 & FitPrune~\cite{ye2025fit} & General & Yes & Minimizes divergence of attention distributions \\
 & ATP-LLaVA~\cite{ye2025atp} & General & No & Learnable module for input-adaptive pruning \\ \cmidrule(l){2-5} 
 & FrameFusion~\cite{fu2024framefusion} & Video & Yes & Pruning in deep layers based on cumulative attention scores \\
 & HoliTom~\cite{shao2025holitom} & Video & Yes & Uses cumulative attention scores for pruning inside LLM \\
 & StreamingVLM~\cite{xu2025streamingvlm} & Streaming Video & No & Keeps attention sinks and aligns training with streaming inference \\ \bottomrule
\end{tabular}%
\caption{Representative prefilling-side token compression methods, categorized by the optimization signal (Diversity or Attention) and input modality (general vision-language, video, and streaming video).}
\label{tab:prefill_token_compress}
\end{table*}

\subsection{Prefilling-Side Token Compression}
\label{ssec:prefilling_token_compression}

Unlike encoding-side compression, prefilling-side strategies operate within the LLM backbone's latent space ($\mathbf{H}_v$). By leveraging cross-modal semantic signals available only after projection, these methods achieve potentially higher compression ratios, directly mitigating the quadratic attention bottleneck during prefilling. As summarized in \Cref{tab:prefill_token_compress}, we categorize techniques by their optimization signal: {diversity-guided} (minimizing redundancy) and {attention-guided} (maximizing saliency).
Within each category, we further classify these methods with the corresponding \emph{input modality}.


\paragraph{Diversity-Guided Compression.} 
These methods operate on the premise that visual tokens exhibit high spatial and temporal correlation~\cite{chen2025variation}. The objective is to retain a subset of tokens that maximizes semantic coverage while minimizing embedding similarity. Techniques like {G-Prune}~\cite{jiang2025kind}, {DivPrune}~\cite{alvar2025divprune}, and {CDPruner}~\cite{zhang2025beyond} utilize clustering algorithms or Determinantal Point Processes to identify and merge redundant tokens based on geometric distance in the feature space.
For videos with temporal dimensions, methods such as {DyCoke}~\cite{tao2025dycoke}, {FastVID}~\cite{shen2025fastvid},{HoliTom}~\cite{shao2025holitom} and VidCom$^2$~\cite{liu2025vidcom2} extend this logic to spatiotemporal merging. They fuse temporally adjacent or spatially similar patches across frames~\cite{lin2026v}, preserving the ``motion flow'' while discarding static redundancies.

\paragraph{Attention-Guided Compression.} 
This paradigm leverages LLMs' intrinsic self-attention weights as a proxy for token utility~\cite{liu2026globalcom2}. {FastV}~\cite{chen2024image} and {PyramidDrop}~\cite{xing2024pyramiddrop} observe that early-layer attention patterns are strong predictors of deep-layer relevance. They employ ``early-exit'' strategies, pruning tokens with low cumulative attention scores in initial layers to save compute in deeper layers. 
Advanced variants like {FitPrune}~\cite{ye2025fit} minimize the divergence between full and pruned attention distributions, while {ATP-LLaVA}~\cite{ye2025atp} introduces learnable gating modules. For video, {StreamingVLM}~\cite{xu2025streamingvlm} utilize ``attention sinks'' of both text and visual tokens to maintain reasoning stability over long contexts without quadratic computation overhead and linear memory growth.

\MySummaryBox{Moving beyond binary choices, effective systems are hybridizing these paradigms, fusing outer-LLM diversity filtering with inner-LLM attention pruning to decouple geometric redundancy from semantic reasoning. 
This integration is crucial for addressing the inherent modal asymmetry of LVLMs, 
necessitating cross-modal signals that aggressively compress redundant visual states while preserving textual fidelity. Consequently, the video domain is rapidly pivoting toward real-time streaming, shifting focus from holistic offline processing to progressive, locality-aware mechanisms capable of handling infinite visual contexts.
}

\subsection{Sparse Attention}
\label{ssec:sparse_attention}

To combat the quadratic complexity of prefilling, sparse attention mechanisms restrict computation to high-salience regions. Early generic approaches, such as XAttention~\cite{xu2025xattention} (antidiagonal block scoring) and SpargeAttn~\cite{zhang2025spargeattn} (two-stage online filtering), impose sparsity patterns derived from standard LLM heuristics. However, these methods often overlook the unique structural properties of visual tokens.
Addressing this, MMInference~\cite{li2025mminference} introduces \emph{modality-aware permutation}, optimizing sparse kernels by explicitly modeling the distinct attention signatures of visual versus textual data. 
For video, Video-XL-2~\cite{qin2025video} introduces chunk-based prefilling that divides visual sequence into chunks where tokens attend only to their local chunk and coarse-grained historical timestamp tokens.
Pushing this further, VideoNSA~\cite{song2025videonsa} shifts from post-hoc masking to \textit{native sparse training}. VideoNSA employs Native Sparse Attention (NSA)~\cite{yuan2025native} for video tokens while retaining dense attention for text to preserve reasoning capability.

\MySummaryBox{The evolution of sparse attention in LVLMs suggests the necessity to consider modality-aware architectures that integrate sparsity objectives directly into the training loop.}
\section{Efficiency Techniques at Decoding}
\label{sec:decoding}

This section surveys efficiency techniques for LVLM decoding, where textual output is generated token-by-token. 
Governed by TPOT ($\tau_{\text{DEC}}$ in \Cref{equation:total_latency}), this stage is strictly memory-bound: latency is dominated by the limited bandwidth $\beta_{\text{mem}}$ required to load model weights $|\psi|$ and the dynamic KV cache $|\mathcal{K}\mathcal{V}|_i$ at each required step $i$.
To address these constraints, we structure the landscape into three strategic axes:
i) \emph{KV cache compression} (\Cref{ssec:kv_compression}) that reduces the memory footprint $|\mathcal{K}\mathcal{V}|_i$, directly alleviating the bandwidth bottleneck; ii) \emph{speculative decoding} (\Cref{ssec:sepculative_decoding}) that breaks the sequential dependency, amortizing the cost of large-model verification over rapid, lightweight draft steps; and iii) \emph{efficient reasoning} (\Cref{ssec:efficient_reasoning}) that targets reducing the generation length $N_{o}$ via optimizing the conciseness of reasoning chains.

\begin{table*}[!t]
\centering
\tiny
\renewcommand{\arraystretch}{1.2}
\begin{tabular}{@{}p{1.5cm} l p{0.8cm} p{9cm}@{}}
\toprule
\textbf{Granularity} & \textbf{Method} & \textbf{Scenario} & \textbf{Key Strategy \& Insight} \\
\midrule
\multirow{6}{*}{\textbf{Token-Level}} 
 & LOOK-M~\cite{wan2024look} & Static & \textbf{Text-Prior Pruning}: Prioritizes textual KVs; evicts visual tokens based on attention scores. \\
 & Elastic Cache~\cite{liu2024efficient} & Static & \textbf{Merging}: Fuses less important KVs guided by distinct encoding/decoding metrics. \\ 
 & FastCache~\cite{zhu2025fastcache} & Serving & \textbf{Self-supervised}: Uses a lightweight modality-specific compressor to reduce overhead. \\
 & Inf-MLLM~\cite{ning2024inf} & Streaming & \textbf{Bias Adjustment}: Maintains compact cache with adjustable attention bias for long-term dependency. \\
 & SparseVILA~\cite{khaki2025sparsevila} & Streaming & \textbf{Decoupled Sparsity}: Decouples query-agnostic pruning (prefill) and query-aware retrieval (decoding). \\ 
 & ReKV~\cite{di2025streaming} & Streaming & \textbf{Retrieval}: Offloads video chunks to external memory and selectively retrieves query-relevant KVs. \\
 & LiveVLM~\cite{ning2025livevlm} & Streaming & \textbf{Dual-Memory}: Combines a short-term sliding window with retrieval from compressed long-term memory. \\
\midrule
\multirow{5}{*}{\textbf{Layer-Level}} 
 & VL-Cache~\cite{tu2024vl} & Static & \textbf{Sparsity-based}: Allocates larger cache budgets to layers with denser attention patterns. \\
 & MEDA~\cite{wan2025meda} & Static & \textbf{Entropy-based}: Guided by cross-modal attention entropy to preserve complex interactions. \\
 & ST3~\cite{zhuang2025st3} & Static & \textbf{Progressive Pruning}: Prunes more visual tokens in deeper layers based on decreasing visual importance. \\
 & MadaKV~\cite{li2025madakv} & Static & \textbf{Inter-layer Compensation}: Adjusts subsequent layer budgets based on current compression. \\
 & InfiniPot-V~\cite{kim2025infinipot} & Streaming & \textbf{Adaptive Pooling}: Uses varying pooling kernel sizes across layers to balance abstraction and detail. \\
\midrule
\textbf{Head-Level} & SparseMM~\cite{wang2025sparsemm} & Static & \textbf{Asymmetric Budget}: Identifies vital visual heads and allocates higher budgets to them. \\
\midrule
\multirow{3}{*}{\textbf{Bit-Level}} 
 & AKVQ-VL~\cite{su2025akvq} & Static & \textbf{Adaptive Mixed-Precision}: High bit-width for critical tokens, 2-bit for others. \\
 & VidKV~\cite{tao2025plug} & Static & \textbf{Sub-2-bit}: Differential treatment for K (channel-wise) and V (1.58-bit + salient token preservation). \\
 & CalibQuant~\cite{han2025calibquant} & Static & \textbf{Calibrated 1-bit}: Channel-wise 1-bit quantization with post-calibration for extreme values. \\
\bottomrule
\end{tabular}%
\caption{Representative KV cache compression methods, categorized by operational granularity (token, layer, head, and bit) and inference scenario (static, streaming, and serving).}
\label{tab:kv_compression_overview}
\end{table*}

\subsection{KV Cache Compression}
\label{ssec:kv_compression}
KV cache compression optimizes $\tau_{\text{DEC}}$ by minimizing the effective number of processed KV pairs~\cite{feng2025tamingfragilitykvcache, feng2025identifycriticalkvcache, feng2025adakvoptimizingkvcache}. 
Unlike generic compression, LVLM-specific methods exploit \emph{modal asymmetry}, the observation that visual tokens exhibit far higher redundancy than textual tokens. As categorized in \Cref{tab:kv_compression_overview}, techniques operate across four granularities:
\textbf{\emph{Token-Level}}: Methods like {LOOK-M}~\cite{wan2024look} and {ReKV}~\cite{di2025streaming} employ post-hoc pruning or retrieval strategies, decoupling the massive prefill context from the active working set by offloading or evicting non-salient visual states.
\textbf{\emph{Layer/Head-Level}}: Methods like {VL-Cache}~\cite{tu2024vl},{SparseMM}~\cite{wang2025sparsemm}, and MixKV~\cite{liu2025mixkv} optimize structural allocation, assigning larger cache budgets to ``dense'' layers or ``heads'' that handle cross-modal reasoning. 
\textbf{\emph{Bit-Level}}: Methods like {VidKV}~\cite{tao2025plug} push the limits of precision, utilizing sub-2-bit quantization for robust visual tokens while preserving precision for sensitive text tokens.



\MySummaryBox{While current methods typically function in isolation, the distinct redundancy profiles of LVLMs demand hybrid frameworks that synergize retrieval (context), pruning (sparsity), and quantization (density).
We conduct a pilot exploration of this unified paradigm in~\Cref{ssec:exp_hybrid_kv_compress}.}

\subsection{Speculative Decoding}
\label{ssec:sepculative_decoding}

Speculative decoding (SD) accelerates inference by decoupling generation into rapid drafting (via a lightweight draft model and parallel verification (via the target model). While effective in LLMs~\citep{xia2025swift,zhang2024draft,DBLP:journals/corr/abs-2505-16162}, LVLMs introduce a unique bottleneck: the \emph{visual memory wall}, where the computational cost of processing massive visual contexts ($|\mathcal{K}\mathcal{V}|_i$) erodes the efficiency gains of the draft model.


Most existing SD adaptations are training-aware, focusing on visually specialized draft models. MSD~\cite{lin2025speculative} and Spec-LLaVA~\cite{huo2025spec} utilize multi-stage training or distillation to align draft capabilities. 
To optimize visual processing, FLASH~\cite{wang2025flash}, ViSpec~\cite{kang2025vispec} and SpecVLM~\cite{huang2025specvlm} introduce mechanisms like semi-autoregressive heads or adaptive visual compression. 
Alternatively, HiViS~\cite{xie2025hivis} and FastVLM~\cite{bajpai2025fastvlm} reduce computational costs by reusing the target model's hidden states or early layers, bypassing raw visual inputs.
To bypass training overhead, training-free SD prioritizes direct deployment. In video scenarios, SpecVLM~\cite{ji2025specvlm} exploits the draft model's insensitivity to visual density and performs visual token pruning for the draft model.



\MySummaryBox{Despite validating SD's potential in LVLMs, existing frameworks suffer from a rigid verification bottleneck. They ignore the inherent semantic flexibility of visual descriptions. We argue that visual tasks admit substantial room for relaxed verification, a hypothesis we empirically validate in \Cref{ssec:exp_loosely_sd}.}


\subsection{Efficient Reasoning}
\label{ssec:efficient_reasoning}

Efficient reasoning targets the output horizon $N_o$, aiming to mitigate the latency cost of Chain-of-Thought (CoT) by dynamically aligning inference depth with problem complexity.
Current strategies rely on adaptive computation length regulation in various multimodal scenarios.
PixelThink~\cite{wang2025pixelthink} leverages reinforcement learning to modulate reasoning length, while FS-VisPR~\cite{li2025adaptive} implements a ``fast-slow'' routing mechanism, dispatching queries between lightweight direct solvers and heavy programmatic workflows. 
Similarly, CAR~\cite{lu2025prolonged} adopts an uncertainty-driven expansion, triggering extended reasoning chains only when initial confidence is low.




\MySummaryBox{While effective, these methods rely on coarse instance-level mechanisms.
New opportunities remain for step-level optimization: strategically pruning steps within a generated chain rather than simply processing the entire one.}

\section{Challenges and Future Directions}
\label{sec:futurework}
We identify three algorithmic frontiers targeting the distinct bottlenecks of representation, generation, and continuity. Crucially, we argue that their ultimate realization hinges on a fourth, integrative trajectory: end-to-end system co-design, which unifies these optimization primitives into a cohesive, hardware-aware deployment paradigm.



\paragraph{Representation: Hybrid Compression.}
Employ a uniform strategy~\cite{wan2024look} or adjusting budget allocation alone~\cite{li2025madakv, wang2025sparsemm} are insufficient for the heterogeneous entropy of LVLMs. 
As preliminarily explored in \Cref{ssec:exp_hybrid_kv_compress}, the frontier may lie in \emph{strategic orchestration}: assigning distinct operators (retrieval, pruning, and quantization) tailored to the specific sensitivity of each component.

\paragraph{Generation: Modality-Aware Decoding.}
To overcome the visual memory wall, current efficient decoding strategies~\cite{xie2025hivis,ji2025specvlm,gao2025aim} must abandon generic NLP heuristics. The path forward requires resolving two deficits: (i) Visual Draft Alignment, ensuring lightweight drafters can handle dense visual contexts, and (ii) Relaxed Verification, moving from rigid exact-match criteria to semantic-aware validation (as supported by \Cref{ssec:exp_loosely_sd}).




\paragraph{Continuity: The Streaming Pivot.}
The transition from offline processing to infinite-context streaming demands a shift from holistic analysis to progressive state management~\cite{xu2025streamingvlm}. Future work should prioritize stage-specific optimizations, such as streaming visual memory management at encoding~\cite{zhang2025flash}, progressive token compression at prefilling~\cite{chen2025streamingtom,xu2025streamingvlm,wang2025accelerating}, and locality-aware KV cache compression at decoding~\cite{ning2025livevlm}. 
Sustaining unbounded throughput will require synergizing training-free heuristics~\cite{chen2025streamingtom} with training-aware paradigms~\cite{xu2025streamingvlm,zhang2025flash} to prevent resource saturation.

\paragraph{The Unifying Imperative: End-to-End System Co-Design.}
%
Algorithm-level optimizations often falter against system-level bottlenecks~\cite{zhang2025hmi} like bandwidth saturation and pipeline bubbles. Emerging disaggregated architectures (e.g., EPDServe~\cite{singh2024efficiently}, ModServe~\cite{qiu2025modserve}) demonstrate the necessity of mapping distinct inference stages to specialized hardware. The critical path forward lies in hardware-algorithm co-design, unifying architectural tailoring with semantic-aware predictive scheduling.
We provide a detailed analysis of these serving architectures and their evaluation standards in~\Cref{app:sys}.

\section{Literature Selection Protocol}

We followed a systematic three-phase protocol to curate the literature included in this survey:

\paragraph{Broad Exploration.}
We began with a broad search on Google Scholar to identify the major research themes, representative architectures, and key terminology related to large vision-language models and efficient inference.

\paragraph{Targeted Filtering.}
Based on the initial candidate pool, we performed targeted screening over papers from major venues in NLP, machine learning, artificial intelligence, and computer vision, including ACL, EMNLP, NAACL, ICML, NeurIPS, ICLR, CVPR, and ICCV, as well as relevant arXiv preprints. We primarily focused on work published from 2020 to early 2026.

\paragraph{Bidirectional Citation Tracking.}
To further improve coverage, we applied bidirectional citation tracking. We traced backward from seminal papers such as LLaVA and BLIP-2 to identify foundational work, and traced forward to capture recent extensions and state-of-the-art systems, including representative models from the Qwen series.

\section{Positioning in the Evolving Landscape}
\label{app:related_survey}

The surge in LVLMs has been accompanied by a proliferation of survey literature focusing on computational efficiency. To clarify the unique contributions of our work, we position this survey within the broader landscape of Large Language Model (LLM) and Multimodal Large Language Model (MLLM) research.

\paragraph{Comparison with LLM-Centric Surveys.}
Existing efficiency research has predominantly focused on the text modality, spanning the spectrum from algorithmic optimizations to system-level serving. Broad-spectrum surveys have systematized these efforts through data-, model-, and system-level perspectives~\citep{ELLMSurvey, wan2024efficientlargelanguagemodels}, with recent comprehensive tutorials further establishing full-stack taxonomies that link algorithmic design directly to hardware bottleneck diagnosis~\cite{Ning2025Tutorial}. Complementing these holistic views, specialized reviews delve into specific techniques like quantization and alternative architectures~\citep{11247933, sun2025speedwinssurveyefficient}, while deployment-centric works emphasize MLSys challenges such as request scheduling and cluster-level load balancing~\citep{zhen-etal-2025-taming, Miao_2025}. 
While these works establish fundamental principles for text generation, they do not address the unique ``visual memory wall'' and the specific pipeline bottlenecks inherent in processing fine-grained visual inputs.

\paragraph{Comparison with MLLM-Centric Surveys.}
Surveys in the multimodal domain typically prioritize different thematic axes. \textit{data-centric perspectives} focus exclusively on data preparation and post-training~\cite{zhang2025train} techniques like synthesis and distillation \cite{bai2024surveymultimodallargelanguage, luo-etal-2025-survey}. \textit{architectural overviews} provide taxonomies of model structures and training recipes \cite{zhang-etal-2024-mm}, often targeting edge computing scenarios \cite{jin2024efficientmultimodallargelanguage,zhang2025sf} or resource-constrained devices \cite{shinde2025surveyefficientvisionlanguagemodels,zhou2025floe}. Finally, \textit{modality-specific} reviews focus narrowly on Vision-Language-Action (VLA) Models  \cite{yu2025surveyefficientvisionlanguageactionmodels} or token compression across images and videos to mitigate quadratic attention \cite{shao2025tokenstalkmuchsurvey}. Unlike these isolated optimizations, we focus on \textit{stage-aware algorithmic optimizations} across the end-to-end inference pipeline.

\paragraph{Unique Contribution: End-to-End LVLMs Inference.}
In contrast to prior reviews that often focus on isolated optimizations, this survey provides a systematic analysis of the end-to-end LVLMs inference pipeline. We distinguish our contribution through three primary dimensions. First, we provide a \textit{stage-specific taxonomy} along three execution stages: \emph{encoding}, \emph{prefilling}, and \emph{decoding}. Second, we conduct a \textit{bottleneck-aware analysis} to examine how overhead is shaped not only by compute but by memory traffic, cache locality, and sequence length, specifically addressing the transition from compute-bound encoders to bandwidth-bound decoding. Third, we offer a synthesis of \textit{design principles and prospects}, identifying the pivotal shift toward dynamic, density-aware mechanisms and advocating for stage-disaggregated serving architectures to guide future research.

\section{Conclusion}
This survey systematizes efficient LVLM inference through a stage-aware taxonomy covering \emph{encoding}, \emph{prefilling}, and \emph{decoding}. We identify the critical bottleneck shift from compute-bound visual encoding to memory-bound autoregression, showing that efficiency hinges on mitigating \emph{visual token dominance} across the pipeline. 
Crucially, our analysis locates the algorithmic frontier in three modality-centric shifts: from uniform compression to hybrid orchestration, from rigid verification to semantic-aware relaxation, and from holistic processing to progressive state management.
Ultimately, we argue that the advancement of this field necessitates a shift from isolated algorithmic enhancements to holistic, full-stack optimizations.

\section{Acknowledgements}
The work was supported by the Major Research Program of the Zhejiang Provincial Natural Science Foundation (Grant No.~LD24F020015), CCF-Baidu Open Fund (No.~202509), and Zhejiang Province "Leading Talent of Technological Innovation Program" (No.~2023R5214).
\clearpage

\section{Limitations}
While this survey synthesizes efficient inference methodologies across encoding, prefilling, and decoding stages, the rapid release of proprietary models (e.g., GPT-4o) means some undocumented, closed-source optimizations may be omitted. Crucially, our analysis prioritizes the massive computational redundancy in image and video scenarios, where the visual memory wall is most acute. Consequently, domain-specific optimizations for document understanding (e.g., layout-driven cropping, OCR-aware patching) and heterogeneous multi-image scheduling receive less depth, as their discrete token structures diverge from the continuous temporal focus of this work. Finally, we concentrate on latency and memory throughput, leaving energy efficiency and theoretical compression bounds for future investigation. We advocate for standardized, hardware-agnostic benchmarks to further guide the deployment of next-generation LVLMs.
\section{Ethical Considerations}
This work synthesizes existing literature and involves no human subjects. The discussed methods aim to advance Green AI by reducing energy consumption and democratizing access to multimodal systems. However, we caution that efficiency-oriented optimizations, particularly lossy compression, pose risks, including the potential degradation of safety guardrails and increased hallucination rates. We urge the community to adopt robust evaluation protocols that monitor these ethical dimensions alongside latency metrics.

Regarding our pilot experiments, all evaluations are conducted using open-source models and datasets in strict compliance with their respective licenses (Apache 2.0 and CC-BY-4.0). We utilize MileBench, VideoChatGPT, and VideoDetailCaption benchmarks to ensure reproducibility.\footnote{Dataset sources: MileBench and VideoChatGPT are under Apache 2.0; VideoDetailCaption is under CC-BY-4.0. Access URLs are available at: \url{https://github.com/MileBench/MileBench}, \url{https://huggingface.co/datasets/lmms-lab/VideoChatGPT}, and \url{https://huggingface.co/datasets/lmms-lab/VideoDetailCaption}.}

\bibliography{custom}

\clearpage
\appendix
\crefalias{section}{appendix}
\crefalias{subsection}{appendix}

\section{Takeaways}
\label{app:takeaway}

Through a comprehensive review of efficiency techniques for LVLM inference, we distill the following critical insights across the three execution stages. 
These takeaways highlight the shifting bottlenecks and the emerging design principles for the next-generation LVLM systems.

\subsection{Efficiency Techniques at Encoding Stage}

The encoding phase dictates the initial computational footprint of the entire inference lifecycle. Decisions made here regarding resolution and feature granularity set the baseline cost for downstream processing. Our analysis identifies a decisive shift from static, one-size-fits-all preprocessing to dynamic, density-aware mechanisms that align computational expenditure with information content.

\begin{takeaway}

\smallskip

\ding{202} \textbf{Encoding as the Primary Lever for Token Efficiency.} Improving token efficiency at the encoding stage (e.g., via pooling or C-Abstractors) offers the highest leverage within the inference pipeline. Removing a single token at this point eliminates its computation across all subsequent LLM layers and prevents it from occupying space in the KV cache. Therefore, the encoding stage is decisive for downstream token efficiency. Future architectures will likely favor end-to-end learnable compressors over heuristic selection to maximize this upstream efficiency.
\smallskip

\ding{203} \textbf{Paradigm Shift from Fixed to Dynamic.} Visual information naturally exhibits non-uniform density and substantial redundancy. Traditional preprocessing methods that treat all visual information uniformly are becoming increasingly obsolete. The SOTA models (e.g.,~Qwen3-VL~\cite{yang2025qwen3} and LLaVA-Next~\cite{liu2024llavanext}) suggest that Native Dynamic Resolution is a promising direction, as it adaptively allocates tokens based on the information density of the input, simultaneously preserving fine-grained visual details and reducing the total token count.
Moreover, dynamic token budgeting across regions of images and frames of videos is becoming increasingly mainstream, further underscoring this trend.
\smallskip

\ding{204} \textbf{Encoding is Compute-Bound, Not Memory-Bound.} Unlike the decoding stage, the visual encoding phase is characteristically compute-bound. As input resolutions scale (e.g., to 4K), the latency of the Vision Transformer (ViT) becomes a significant component of the (TTFT). Consequently, optimizations that target architectural efficiency, such as structural reparameterization (e.g., FastViT~\cite{vasu2023fastvit}) and operator fusion, yield higher end-to-end acceleration gains than simple weight quantization in this stage.
\end{takeaway}

\subsection{Efficiency Techniques at Prefilling Stage}

Optimizing the prefilling stage is fundamentally about mitigating the quadratic scaling of attention mechanisms ($\mathcal{O}(N^2)$) in the face of increasingly long visual contexts. As models scale to handle high-resolution imagery and long-form video, the latency of the first token (TTFT) becomes a primary bottleneck. The literature converges on the insight that visual data exhibits significantly higher redundancy than text, permitting aggressive, non-uniform compression strategies.

\begin{takeaway}

\smallskip

\ding{202} \textbf{The Trend toward Hybrid Token Compression Paradigms.}
Emerging trends in prefilling-side token compression indicate a shift from isolated processing toward hybrid, multi-stage architectures (e.g., FrameFusion~\cite{fu2024framefusion}, Holitom~\cite{shao2025holitom}) that synergize diversity-guided and attention-guided mechanisms.
This paradigm leverages a "coarse-to-fine" strategy: maximizing compression rates by first eliminating intrinsic visual redundancy via token diversity outside the LLM backbone, and subsequently refining the selection based on attention distribution inside the LLM backbone. 
\smallskip

\ding{203} \textbf{Asymmetric Modality Redundancy.} Visual tokens exhibit significantly higher spatiotemporal redundancy compared to the semantic redundancy of text tokens. Uniform pruning or sparse attention strategies often fail because they treat both modalities equally. Optimal prefilling requires modality-aware policies that apply aggressive compression~\cite{zhang2024sparsevlm} and transformation~\cite{li2025mminference} to visual tokens (based on similarity or attention) while maintaining a conservative approach for text tokens to prevent semantic collapse.
\smallskip

\ding{204} \textbf{The Trajectory toward Real-time Streaming Videos.} 
Research is pivoting from offline video understanding (e.g., DyCoke~\cite{tao2025dycoke}, VidCom2~\cite{liu2025video})—characterized by static, fully known contexts—to online streaming paradigms (e.g., StreamingTOM~\cite{chen2025streamingtom}, StreamingVLM~\cite{xu2025streamingvlm}). Streaming scenarios require processing infinite visual sequences with minimal delay, redefining prefilling from a singular initialization event to a recurring operational cost. Technical approaches transition from compression based on holistic video information toward progressive and locality-aware compression mechanisms to effectively minimize the recurring prefill costs.

\end{takeaway}

\subsection{Efficiency Techniques at Decoding Stage}

The decoding stage is characteristically memory-bound, defined by the operational intensity of loading massive Key-Value (KV) caches for autoregressive generation. In LVLMs, this bottleneck manifests as a ``Visual Memory Wall'': for long-context multimodal inference, the KV cache footprint often exceeds model weights themselves, with visual tokens accounting for 80\%-90\% of the total memory usage. However, the generation phase relies predominantly on textual history and only sparse visual cues. The takeaways below distill the emerging design principles that exploit this asymmetry to maximize throughput and minimize latency.

\begin{takeaway}

\smallskip

\ding{202} \textbf{From Coarse to Multi-Granular Modality-Aware Compression.} 
Traditional uniform compression is suboptimal for LVLMs as it ignores the extreme redundancy of visual data and the importance of textual history. The current trend is modality-adaptive compression that operates at finer granularities from layers and heads to bits. Techniques now allocate distinct budgets: applying aggressive compression to visual tokens (e.g., sub-2-bit quantization~\cite{tao2025plug}, sparse head pruning~\cite{wang2025sparsemm}) while retaining high precision for most text tokens. This asymmetric treatment reduces the ``visual memory wall'' payload without compromising much performance.
\smallskip

\ding{203} \textbf{Draft Model Adaptation for Multimodal Speculation.} Applying generic speculative decoding to LVLMs is inefficient if the draft model is burdened by heavy visual processing. The trend is shifting towards visual-lightweight drafting, achieved via two main paths: training specialized lightweight draft models (e.g., MSD~\cite{lin2025speculative}, Spec-LLaVA~\cite{huo2025spec}) or pruning redundant visual tokens from the draft input in a training-free manner (e.g., SpecVLM~\cite{ji2025specvlm}). Both approaches leverage language priors to accelerate candidate generation, effectively amortizing the high bandwidth cost of the target model's verification step.
\smallskip

\ding{204} \textbf{From Monolithic to Disaggregated Serving Architectures.} The resource demands of LVLM inference are highly heterogeneous: Encoding is compute-intensive, while Decoding is bandwidth-intensive. Monolithic serving architectures struggle to balance these conflicting needs. The field is moving toward Stage-Disaggregated Serving (e.g. ModServe~\cite{qiu2025modserve}), where encoding and decoding are decoupled and scheduled onto specialized hardware configurations to maximize cluster-wide utilization and throughput.
\end{takeaway}

\subsection{Efficiency Techniques at the System Level}

Complementing our granular analysis of the encoding, prefilling, and decoding stages, we extend our scope to the holistic serving ecosystem. In \Cref{ssec:system_architecture}, we survey the architectural landscape of efficient LVLM serving systems adopted by the community. By synthesizing the trade-offs identified across these isolated stages with broader system-level constraints, we distill the following key takeaways for optimization.

\begin{takeaway}

\smallskip

\ding{202} \textbf{Resource Decoupling via Spatial Multiplexing.}
Fundamentally, efficient LVLM serving relies on \textit{spatial multiplexing}~\cite{zhangspaceserve}, which decouples inference components onto specialized resource groups. This separation allows each group to be independently configured and scaled, effectively resolving the conflicting resource affinities inherent to distinct inference stages and modalities.
\smallskip

\ding{203} \textbf{Optimal Architecture Selection.}
As synthesized in~\Cref{tab:serving_tradeoff}, the optimal topology is dictated by the workload shape:
(1) \textbf{Stage-based} disaggregation isolates computational bursts, making it ideal for \textit{Long-Video} tasks to stabilize P99 latency.
(2) \textbf{Modality-based} partitioning simplifies operations while maximizing throughput for \textit{Balanced} workloads.
(3) \textbf{Resource-based} multiplexing eliminates communication overhead, serving as the preferred solution for \textit{Latency-Critical} or \textit{Edge} applications~\cite{jin2023emsassist}.
\end{takeaway}

\section{Model Architecture}
\label{app:architecture}
\subsection{Overview}
As formalized in~\Cref{ssec:canonical_arch}, modern LVLMs converge on a unified three-component architecture: vision encoder $\mathcal{E}_\phi$, modality adapter $\mathcal{A}_\theta$, and LLM backbone $\mathcal{L}_\psi$. This appendix provides implementation details and a taxonomy of representative models organized by their efficiency-oriented design choices.

We detail the implementation variants of each component and categorize representative models by their efficiency strategies.

\subsection{Core Components}
\paragraph{Vision Encoder.}
The vision encoder $\mathcal{E}_\phi$ produces patch embeddings $\mathbf{X}_v \in \mathbb{R}^{N_p \times D_v}$ from raw visual input.
Modern LVLMs typically reuse pretrained visual encoders such as CLIP~\cite{radford2021learning}, MetaCLIP~\cite{xu2023metaclip}, EVA-CLIP~\cite{sun2023eva}, SigLIP~\cite{zhai2023sigmoid}, or ViT~\cite{DBLP:conf/iclr/DosovitskiyB0WZ21} as general-purpose front ends. These encoders underpin many widely used systems. For instance, LLaVA~\cite{liu2023llava} and LLaVA-OneVision~\cite{an2025llavaonevision15} leverage CLIP and SigLIP variants, respectively, while InternVL~\cite{chen2024internvl} and Qwen-VL~\cite{bai2023qwen} utilize scale-up strategies based on powerful backbones such as InternViT and OpenCLIP. They convert images into patch-based token sequences whose granularity and resolution define the initial size of the multimodal context.

Video-centric LVLMs extend this paradigm to the temporal dimension.
Early approaches like Video-ChatGPT~\cite{maaz2024video} apply average pooling over frame-level features to obtain compact representations. In contrast, models like Video-LLaMA~\cite{zhang2023video} and VideoChat~\cite{li2023videochat} utilize a Video Q-Former to aggregate temporal information. More recent efficient models, such as VideoLLaMA~3~\cite{zhang2025videollama3frontiermultimodal} and SlowFast-LLaVA~\cite{xu2024slowfastllavastrongtrainingfreebaseline}, employ hierarchical or dual-stream encoders to capture spatiotemporal dependencies without explicitly expanding the token count linearly with frame numbers.

Across these models, the vision encoder controls the number and density of visual tokens generated during the encoding stage and is therefore one of the primary factors shaping computational and memory cost throughout the inference pipeline.

\paragraph{Modality Adapter.}
The modality adapter $\mathcal{A}_\theta$ maps $\mathbf{X}_v$ to visual context $\mathbf{H}_v \in \mathbb{R}^{N_v \times D_{\mathcal{L}}}$. Modern implementations fall into two categories:
\begin{enumerate}
\item \textbf{Linear Projection.} Exemplified by LLaVA-1.5~\cite{liu2023improvedllava} and InternVL-3.5~\cite{wang2025internvl35}, this approach uses a simple MLP with compression ratio $r = N_v/N_p = 1$, preserving full visual granularity but incurring high prefilling cost.

\item \textbf{Learnable Query-Based Mechanisms.}
Pioneered by models like BLIP-2~\cite{li2023blip} and Video-LLaMA~\cite{zhang2023video}, these methods utilize a fixed set of latent queries (e.g., via Q-Former or Video Q-Former) to extract semantic information from variable-length visual features.
This process compresses dense visual inputs into a compact, fixed-length sequence of tokens regardless of the input resolution.
Recent works such as Dynamic-VLM~\cite{wang2025dynamic} and TokenPacker~\cite{li2025tokenpacker} further refine this paradigm by introducing dynamic compression rates or coarse-to-fine injection schemes, aiming to balance high compression ratios with the preservation of fine-grained spatial details.
\end{enumerate}

\paragraph{LLM Backbone.}
The LLM backbone $\mathcal{L}_\psi$ processes joint context $\mathbf{C}$ (concatenation of visual and text embeddings) to generate responses. Modern implementations build on pretrained LLMs: server-scale backbones like LLaMA~\cite{grattafiori2024llama3herdmodels}, Qwen~\cite{yang2025qwen3}, Mistral~\cite{jiang2023mistral7b}, and InternLM~\cite{cai2024internlm2technicalreport}, or lightweight variants like Phi~\cite{abdin2024phi3technicalreporthighly} and Gemma~\cite{gemmateam2024gemmaopenmodelsbased} for edge deployment.

LVLMs differ in how visual tokens are introduced into the backbone:
\begin{enumerate}
\item \textbf{Input Concatenation}: The dominant strategy, pioneered by LLaVA~\cite{liu2023llava}, projects visual tokens into the textual embedding space and concatenates them directly with text tokens at the input layer. This allows visual information to flow through all self-attention layers, enabling deep multimodal interaction. Due to its architectural simplicity and training efficiency, this approach has become the mainstream strategy for recent open-source models.

\item \textbf{Cross-Attention Injection}: In contrast, architectures like LLaMA 3.2-Vision~\cite{grattafiori2024llama3herdmodels} and Flamingo~\cite{alayrac2022flamingovisuallanguagemodel} inject visual information into intermediate layers via interleaved cross-attention modules. 
This approach typically keeps the pretrained LLM parameters frozen (or partially frozen) and uses these adapter layers to fuse visual features conditionally.
While this avoids extending the input context length with dense visual tokens, it necessitates architectural modifications to the attention blocks and introduces additional parameters.
\end{enumerate}

All multimodal reasoning ultimately occurs inside the backbone, its internal pathways determine how tokens are preserved, abstracted, or attenuated as computation proceeds. As a result, many inference-time efficiency techniques operate directly on this module, making it the central substrate governing both capability and efficiency in modern LVLMs.

\subsection{Model Taxonomy}
Although modern LVLMs broadly follow the unified architecture outlined above, existing research exhibits clear differentiation in how visual information is represented, injected, and managed. From an efficiency-oriented perspective, we categorize current LVLMs into three groups based on their approach to managing visual token count $N_v$ and inference complexity:

\paragraph{Performance-Prioritized Models.}
These models aim to maximize multimodal capability. They primarily concentrate on designing refined training pipelines and curating high-quality training data to ensure robust multimodal alignment. Representative models include InternVL-3.5~\cite{wang2025internvl35}, Qwen2.5-VL~\cite{wang2024qwen2}, DeepSeek-VL2~\cite{wu2024deepseekvl2}, LLaVA-OneVision~\cite{lillava}, Llama-3.2-Vision~\cite{grattafiori2024llama3herdmodels}, CogVLM2~\cite{wang2024cogvlm}, Cambrian-1~\cite{tong2024cambrian1}, Yi-VL~\cite{ai2025yiopenfoundationmodels}, InternLM-XComposer2~\cite{dong2024internlmxcomposer2} and Ovis~\cite{lu2024ovisstructuralembeddingalignment}. These models typically generate dense visual token sequences to preserve fine-grained details, serving as high-computation baselines for efficiency studies.

\paragraph{Partially-Optimized Models.}
These models preserve the high-performance backbone of standard LVLMs but introduce specific optimization strategies targeting isolated bottlenecks, particularly in the adapter or token selection modules.
They strive to balance performance and efficiency by reducing $N_v$ through adapter-level compression or token selection strategies.
Representative models include BLIP-2~\cite{li2023blip}, InstructBLIP~\cite{dai2023instructblip}, Qwen-VL~\cite{bai2023qwen}, MiniGPT-4~\cite{zhu2023minigpt4}, MiniGPT-v2~\cite{chen2023minigptv2}, mPLUG-Owl2~\cite{ye2023mplugowl}, and Honeybee~\cite{cha2024honeybee}.






\paragraph{Holistically-Optimized Models.}
Holistic models aim for end-to-end efficiency through system-level co-design or efficiency-native architectural innovations. A prime example of full-pipeline optimization is NVILA~\cite{liu2025nvila}, which introduces a ``scale-then-compress'' paradigm. It jointly optimizes the architecture by scaling up resolutions for precision while compressing visual tokens for efficiency, and further enhances the entire lifecycle from training to deployment with system-level accelerations. Other works achieve holistic efficiency by redesigning the architecture for specific constraints. MobileVLM V2~\cite{chu2024mobilevlmv2} co-designs a mobile-friendly vision encoder with a tailored small-scale LLM to achieve efficient inference on edge devices.  In the video domain, VideoChat-Flash~\cite{li2025videochatflashhierarchicalcompressionlongcontext} implements a full-pipeline hierarchical compression strategy, progressively reducing redundancy from the visual encoder to the LLM to handle long contexts efficiently.

\paragraph{Discussion of Representative Architectures.}
Representative multimodal architectures also reflect several distinct pathways toward efficient inference. The Qwen series~\citep{bai2025qwen3vltechnicalreport,qwen3.5} exemplifies the shift toward Native Dynamic Resolution, while Qwen3.5 further integrates Hybrid Attention with Sparse MoE to substantially improve long-context efficiency, reportedly achieving up to 19$\times$ higher throughput. DeepSeek-VL2~\cite{wu2024deepseekvl2}, by contrast, serves as a representative sparse-computation architecture whose Mixture-of-Experts (MoE) design effectively decouples overall model capacity from per-token inference cost. InternVL-3.5~\cite{wang2025internvl35} highlights a system-oriented optimization perspective, where the combination of a Visual Resolution Router and Decoupled Deployment yields a 4.05$\times$ system-level inference speedup. Meanwhile, Llama-3.2-Vision~\cite{grattafiori2024llama3herdmodels} adopts Cross-Attention Injection as a memory-efficient alternative to direct visual-text token concatenation, specifically aiming to alleviate the visual memory wall.



\begin{table*}[!ht]
\centering
\setlength{\tabcolsep}{5pt}
\resizebox{\textwidth}{!}{%
\begin{tabular}{@{}l l l c c l@{}}
\toprule
\multirow{2}{*}{\textbf{Paradigm}} & \multirow{2}{*}{\textbf{Optimal Workload}} & \multirow{2}{*}{\textbf{Key Mechanism}} & \multicolumn{2}{c}{\textbf{Performance Impact}} & \multirow{2}{*}{\textbf{Communication Cost}} \\ \cmidrule(lr){4-5} 
 &  &  & \textbf{P99 TTFT} & \textbf{TPOT} &  \\ \midrule
\textbf{Stage-based} & \textbf{Long Video / Heavy Prefill} & Inter-device Disaggregation & $\downarrow\downarrow$ & $\uparrow$ & \textbf{High} \\
\textit{(e.g., EPDServe)} & (Prefill-dominant) & (Avoids Compute Blocking) & (Best Stability) & (Dedicated) & (Context/KV Transfer) \\ \addlinespace
\textbf{Modality-based} & \textbf{Balanced Multimodal} & Inter-device Partitioning & $-$ & $\uparrow\uparrow$ & \textbf{Medium} \\
\textit{(e.g., ModServe)} & (General VQA) & (Resource Specialization) & (Stable) & (Pipeline) & (Embedding Transfer) \\ \addlinespace
\textbf{Resource-based} & \textbf{Latency-Sensitive / Edge} & SM-level Multiplexing & $\downarrow$ & $\uparrow$ & \textbf{None} \\
\textit{(e.g., SpaceServe)} & (Real-time Interaction) & (Zero Network Hops) & (Fastest Avg.) & (Utilization) & (Intra-GPU Fusion) \\ \bottomrule
\end{tabular}
}
\caption{Qualitative comparison of LVLM serving paradigms, mapping architectural choices to workload characteristics. Legend: $\downarrow$/$\uparrow$ denotes latency/throughput improvement; P99 TTFT indicates worst-case stability.}
\label{tab:serving_tradeoff}
\end{table*}

\begin{table*}[!ht]
    \centering
    \small
    \begin{tabular}{@{}llp{2.6cm}p{8.6cm}@{}}
        \toprule
        \textbf{Metric} & \textbf{Category} & \textbf{Formulation} & \textbf{Definition} \\
        \midrule
        \textbf{TTFT} & Latency 
        & $t_{\text{first}} - t_{\text{arr}}$ 
        & \textit{Time to First Token}. The duration of the prefilling phase, measured from the request arrival time $t_{\text{arr}}$ to the first token generation $t_{\text{first}}$. \\
        \textbf{TPOT} & Latency 
        & $\frac{t_{\text{end}} - t_{\text{first}}}{N}$ 
        & \textit{Time Per Output Token}. The generation speed during decoding, measured from the first token $t_{\text{first}}$ to the last token generation $t_{\text{end}}$, averaged over $N$ tokens. \\
        \textbf{SLO Attainment} & Reliability 
        & $\frac{1}{M} \sum_{i=1}^{M} \mathbb{I}(L_r \le \tau)$ 
        & \textit{Service Level Objective Attainment}.
        Proportion of total requests $M$ where the end-to-end latency $L_r$ of request $r$ satisfies the threshold $\tau$. $\mathbb{I}(\cdot)$ is the indicator function.\\
        \textbf{Goodput} & Throughput 
        & $\frac{1}{T} \sum_{i=1}^{M} \mathbb{I}(L_r \le \tau)$ 
        & \textit{Effective Throughput Rate}. The number of requests per second that strictly satisfy the SLO constraint $\tau$, calculated over the total serving duration $T$. \\
        \bottomrule
    \end{tabular}
    \caption{Taxonomy of efficiency metrics, categorized by performance dimension (Latency, Reliability, Throughput).}
    \label{tab:metrics}
\end{table*}

\begin{table*}[!ht]
    \centering
    \small
    \begin{tabular}{@{}llp{10.5cm}@{}}
        \toprule
        \textbf{Domain} & \textbf{Task Competency} & \textbf{Representative Benchmarks} \\
        \midrule
        & Multimodal Reasoning & MathVista~\cite{lu2023mathvista}, MMMU~\cite{yue2024mmmu}, MathVision~\cite{wang2024measuring}, DynaMath~\cite{zou2024dynamath}, LogicVista~\cite{xiao2024logicvista}, VPCT~\cite{shao2024visual}, MMMU-Pro~\cite{yue2025mmmu}, EMMA~\cite{hao2025can}, SFE~\cite{zhou2025scientists}, ZeroBench~\cite{roberts2025zerobench}, WebQA\cite{chang2022webqa}, MultiModalQA\cite{talmor2021multimodalqa} \\
        \cmidrule(l){2-3}
        \textbf{Image} & General Visual QA & HallusionBench~\cite{guan2024hallusionbench}, MMStar~\cite{chen2024we}, MMBench~\cite{liu2024mmbench}, MUIRBench~\cite{wang2024muirbench}, MMVP~\cite{zhang2024mmvp}, VLMsAreBlind~\cite{rahmanzadehgervi2024vision}, VLMsAreBiased~\cite{vo2025vision}, SimpleVQA~\cite{cheng2025simplevqa}, MME-CC~\cite{zhang2025mme}, MMCoQA\cite{li2022mmcoqa}, SlideVQA\cite{tanaka2023slidevqa} \\
        \cmidrule(l){2-3}
        & Long-Context Understanding & DUDE~\cite{van2023document}, MMLongBench~\cite{wang2025mmlongbench}, OminiDocBench~\cite{ouyang2025omnidocbench}, MileBench\cite{song2024milebench} \\
        \midrule
        & Long Video Understanding & CGBench~\cite{chen2024cg}, LongVideoBench~\cite{wu2024longvideobench}, MLVU~\cite{zhou2025mlvu}, LVBench~\cite{wang2025lvbench}, ALLVB~\cite{tan2025allvb}, VideoMME~\cite{fu2025video},
        VDC~\cite{chaiauroracap} \\ \cmidrule(l){2-3}
        \textbf{Video} & Knowledge \& Reasoning & VideoMMMU~\cite{hu2025video}, MMVU~\cite{zhao2025mmvu}, VCRBench~\cite{qi2025vcr}, VideoReasonBench~\cite{liu2025videoreasonbench}, VideoHolmes~\cite{cheng2025video}, Minerva~\cite{nagrani2025minerva}, VideoSimpleQA~\cite{cao2025video}, NExT-QA\cite{xiao2021next}, Video-ChatGPT\cite{maaz2024video} \\
        \cmidrule(l){2-3}
        & Motion \& Perception & Countix~\cite{dwibedi2020counting}, TVBench~\cite{cores2024tvbench}, TempCompass~\cite{liu2024tempcompass}, TOMATO~\cite{shangguan2024tomato}, MVBench~\cite{li2024mvbench}, EgoTempo~\cite{plizzari2025omnia}, MotionBench~\cite{hong2025motionbench}, Vatex\cite{wang2019vatex} \\
        \bottomrule
    \end{tabular}
    \caption{Taxonomy of LVLM capability benchmarks, categorized by input domain (Image or Video) and chronological evolution from single-image perception to complex reasoning.}
    \label{tab:benchmark_taxonomy}
\end{table*}

\section{System \& Evaluation}
\label{app:sys}
This section surveys the architectural landscape of efficient LVLM serving and outlines the evaluation standards adopted by the community. We first categorize SOTA serving systems based on their decoupling paradigms. Subsequently, we systematize the evaluation landscape by compiling the industry-standard metrics and capability benchmarks commonly used to assess these systems.

\subsection{System Architecture}
\label{ssec:system_architecture}

The core challenge in serving LVLMs stems from the \textit{conflicting resource affinities} inherent to distinct inference phases. The pipeline comprises three stages: Encoding (\texttt{E}), Prefilling (\texttt{P}), and Decoding (\texttt{D}), each exhibiting fundamentally conflicting resource requirements and performance characteristics. Specifically, the \texttt{E} and \texttt{P} stages are compute-intensive with low batch saturation, while the \texttt{D} stage is memory-intensive but supports high batch saturation. This fundamental conflict renders a monolithic, integrated service architecture inefficient. Depending on the specific decoupling granularity and resource allocation strategy, these serving systems can be categorized into three types: \textit{Stage-based}, \textit{Modality-based}, and \textit{Resource-based}.

\paragraph{Stage-based.} Stage-based strategies decompose model inference into distinct temporal stages. EPDServe~\cite{singh2024efficiently} exemplifies this approach by using a black-box optimizer to identify the optimal configuration based on historical workload analysis and introduces dynamic role switching to enhance system adaptivity. RServe~\cite{guo2025rserve} adopts a fine-grained scheduling method to overlap \texttt{E} and \texttt{P} stages, thereby mitigating inter- and intra-request pipeline bubbles. HydraInfer~\cite{dong2025hydrainfer} treats each stage as a composable object (e.g., \texttt{EP+D}, \texttt{E+P+D}) and utilizes a hybrid EPD disaggregation profiler to dynamically deploy the optimal decoupling topology according to Service Level Objective (SLO) requirements.

\paragraph{Modality-based.} These strategies partition resource groups according to the type of request or functional module. ModServe~\cite{qiu2025modserve} segregates resources into a dedicated image pool for visual encoding and a text pool for the LLM backbone. ElasticMM~\cite{liu2025elasticmm} similarly groups instances into text-only and multi-modal pools. However, unlike functional decoupling, ElasticMM executes the entire inference pipeline within each respective group and introduces elastic partition scheduling to dynamically reallocate instances or preempt decoding tasks for prefill bursts based on a gain-cost model. Both approaches rely on inter-device communication to coordinate the multi-modal data flow.

\paragraph{Resource-based.} Unlike inter-device partitioning, this paradigm works at the hardware-resource granularity. SpaceServe~\cite{zhangspaceserve} is a representative example via Streaming Multiprocessor (SM)-level partitioning. It logically separates modality encoders and the text decoder, yet runs them on the same GPU. By assigning SM resources to different tasks, it improves utilization and removes inter-device communication overhead that can bottleneck cross-node modality pools.

Our analysis of the trade-offs across different serving architectures is summarized in \Cref{tab:serving_tradeoff}.

\subsection{Evaluation Standards}
\label{app:taxonomy_evaluation}
To provide a structured understanding of the efficiency landscape, we systematize the evaluation standards into two complementary dimensions: \textit{efficiency metrics} and \textit{capability benchmarks}. This taxonomy serves as a guideline for analyzing the trade-offs between serving latency and multimodal generation quality. Validating optimization techniques requires a dual approach: quantifying speed-up gains using industry-standard metrics while ensuring, through rigorous benchmarking, that model utility is preserved across diverse contexts.

\paragraph{Efficiency Metrics.}
We categorize the metrics used to quantify inference efficiency into latency-oriented and throughput-oriented indicators, as defined in~\Cref{tab:metrics}. These metrics constitute the standard framework for evaluating serving system performance in production environments.

\paragraph{Real-world Alignment: Cost, Scalability, and Energy Efficiency.}
{\color{black}
To bridge the gap between technical metrics and production deployment, we formalize the transition of the metrics defined in~\Cref{tab:metrics} into economic and operational indicators:

\begin{itemize}[itemsep=0pt, topsep=0pt, leftmargin=10pt]
    \item \textbf{Cost Savings (CPSR):} For service providers, raw throughput must be translated into the \textit{Cost Per Successful Request} (CPSR, [\texttt{USD} / req]), representing unit economic efficiency. We derive this as:
    \begin{equation}
        CPSR = \frac{C_{\text{node}}}{G \cdot T_{\text{ref}}}
    \end{equation}
    where $C_{\text{node}}$ denotes the operational expenditure (OpEx) of the hardware instance [\texttt{USD} / node] over a reference time window $T_{\text{ref}}$ [s], and $G$ represents the \textbf{Goodput} [req / s]. By maximizing Goodput, developers can increase request density per hardware unit, directly lowering the CPSR by reducing amortized infrastructure overhead.

    \item \textbf{Scalability Efficiency ($\eta_{\text{scale}}$):} We define scalability as the system's capacity to maintain \textbf{SLO Attainment} under an \textit{iso-resource scaling} scenario, where both hardware capacity $H$ (e.g., number of GPU nodes) and request volume $M$ are increased by a factor of $k$:
    \begin{equation}
        \eta_{\text{scale}} = \frac{\text{SLO Attainment}(k \cdot M \mid k \cdot H)}{\text{SLO Attainment}(M \mid H)}
    \end{equation}
    The notation $(M \mid H)$ denotes the performance measured under workload $M$ conditioned on hardware resources $H$. In large-scale clusters, an ideal system maintains $\eta_{\text{scale}} \approx 1$, signifying linear scalability. A significant degradation indicates systemic bottlenecks, such as VRAM saturation or interconnect contention, triggering the need for elastic resource orchestration.

    \item \textbf{Energy Efficiency (EPT):} Sustainability is quantified via \textit{Energy Per Token} (EPT, [J / token]). While \textbf{TTFT} captures the compute-bound energy burst during prefilling, cumulative energy is primarily governed by \textbf{TPOT}. Given the low arithmetic intensity of decoding ($AI \ll 1$), EPT is formulated as:
    \begin{equation}
        EPT \approx P_{\text{TDP}} \times TPOT_{\text{effective}}
    \end{equation}
    where $P_{\text{TDP}}$ is the Thermal Design Power [W, or J/s] and $TPOT_{\text{effective}}$ is the effective decoding latency [s / token]. Reducing TPOT minimizes the duration GPUs spend in high-power active states, serving as the primary driver for energy sustainability.
\end{itemize}
}

\paragraph{Capability Benchmarks.}
Optimization strategies must be validated against established performance standards to ensure that efficiency gains do not compromise model fidelity. In \Cref{tab:benchmark_taxonomy}, we curate a taxonomy of prevailing benchmarks spanning static image reasoning and dynamic video understanding. We order these datasets chronologically to illustrate the shift of the community towards increasingly complex tasks, ranging from fine-grained visual perception to long-context temporal reasoning.

\section{Future-Forward Pilot Exploration}
\subsection{Hybrid KV Cache Compression}
\label{ssec:exp_hybrid_kv_compress}
To validate the potential of \textit{hybrid compression mechanisms}, we explore a differentiated strategy that moves beyond uniform compression with adaptation to budget allocation~\cite{zeng2026hybridkvhybridkvcache}. Specifically, we utilize text-visual information to categorize attention heads, thereby allocating varying budgets and orchestrating a hybrid compression mechanism combining pruning and retrieval. We conduct preliminary experiments to evaluate this approach on Qwen2.5-VL-7B~\footnote{\url{https://huggingface.co/Qwen/Qwen2.5-VL-7B-Instruct}}~\cite{bai2025qwen2} using NVIDIA L40S GPUs.

\paragraph{Performance Evaluation.} We compare our hybrid scheme against SOTA KV compression methods in LVLMs (LOOK-M~\cite{wan2024look}, MadaKV~\cite{li2025madakv}, SparseMM~\cite{wang2025sparsemm}) across four image benchmarks (SlideVQA~\cite{tanaka2023slidevqa}, MMCoQA~\cite{li2022mmcoqa}, WebQA~\cite{chang2022webqa}, MultiModalQA (MM-QA)~\cite{talmor2021multimodalqa}) from Milebench and one video benchmark (Video-ChatGPT~\cite{maaz2024video}). Table~\ref{tab:hybrid_perf} shows that the hybrid compression consistently outperforms uniform compression baselines, achieving performance competitive with the full cache upper bound.

\begin{table}[t]
\centering
\resizebox{\columnwidth}{!}{%
\begin{tabular}{@{}l|cccc|ccccc@{}}
\toprule
\multirow{2}{*}{\textbf{Method}} & \multicolumn{4}{c|}{\textbf{Image}} & \multicolumn{5}{c}{\textbf{Video-ChatGPT}} \\
 & \textbf{SlideVQA} & \textbf{MMCoQA} & \textbf{WebQA} & \textbf{MM-QA} & \textbf{CI} & \textbf{DO} & \textbf{CU} & \textbf{TU} & \textbf{CO} \\ \midrule
\rowcolor{gray!10} \textbf{Full Cache} & 83.50 & 66.50 & 76.50 & 75.00 & 3.06 & 3.15 & 3.52 & 2.24 & 2.69 \\ \midrule
\textbf{LOOK-M} & 82.50 & 52.50 & 71.00 & 73.50 & 2.92 & 2.97 & 3.38 & 2.05 & 2.49 \\
\textbf{MadaKV} & 82.00 & 55.00 & 70.50 & 74.50 & 2.94 & 3.03 & 3.41 & 2.02 & 2.56 \\
\textbf{SparseMM} & 83.50 & 62.00 & 70.50 & 75.50 & 2.90 & 2.95 & 3.37 & 1.99 & 2.52 \\
\textbf{Hybrid} & \textbf{83.50} & \textbf{63.00} & \textbf{76.00} & \textbf{76.00} & \textbf{2.99} & \textbf{3.05} & \textbf{3.47} & \textbf{2.15} & \textbf{2.57} \\ \bottomrule
\end{tabular}%
}
\caption{Performance of four KV cache compression on Qwen2.5-VL-7B across image and video tasks. Image tasks use \textit{exact match accuracy}. For Video-ChatGPT, scores (ranging from 0 to 5) are generated by \texttt{gpt-4o-mini} across five dimensions: Correctness of Information (CI), Detail Orientation (DO), Contextual Understanding (CU), Temporal Understanding (TU), and Consistency (CO).}
\label{tab:hybrid_perf}
\vspace{-1em}
\end{table}

\paragraph{Efficiency Evaluation.}Further, we assess the efficiency of the hybrid KV compression with Video-ChatGPT for real-world long video understanding scenarios. We randomly sample 20 data entries and set the maximum generation length to 128 tokens for evaluation. All experiments use FlashAttention~\cite{dao2023flashattention}. As shown in Table~\ref{tab:hybrid_latency}, our method markedly reduces both GPU memory and decoding latency relative to the Full Cache baseline.

\begin{table}[t]
\centering
\resizebox{\columnwidth}{!}{%
\begin{tabular}{@{}lcccc@{}}
\toprule
\multirow{2}{*}{Method} & \multirow{2}{*}{Budget} & Accuracy & GPU Memory & Latency \\
 & & (Avg.) & (GB) & (ms/token) \\ \midrule
\rowcolor{gray!10} \textbf{Full Cache} & 100\% & 2.93 & 1.73 & 58.94 \\
\textbf{Hybrid} & 20\% & \textbf{2.88} & 0.40 & 42.08 \\
\textbf{Hybrid} & 10\% & 2.85 & \textbf{0.22} & \textbf{38.65} \\ \bottomrule
\end{tabular}%
}
\caption{KV cache GPU memory usage and decoding latency on Qwen2.5-VL-7B with Video-ChatGPT.}
\label{tab:hybrid_latency}
\vspace{-1em}
\end{table}

\begin{table*}[!t]
\centering
\small
\resizebox{\textwidth}{!}{
\begin{tabular}{@{}l|l|ccccccccccc@{}}
\toprule
\multirow{3}{*}{\textbf{Target / Draft Model}} &
\multirow{3}{*}{\textbf{Method}} &
\multicolumn{11}{c}{\textbf{Video Detail Caption}} \\
\cmidrule(lr){3-13}
& &
\multicolumn{2}{c|}{\textbf{Main Object}} &
\multicolumn{2}{c|}{\textbf{Detail}} &
\multicolumn{2}{c|}{\textbf{Camera}} &
\multicolumn{2}{c|}{\textbf{Background}} &
\multicolumn{3}{c}{\textbf{Average}} \\
& &
\textbf{Acc(\%)} & \multicolumn{1}{c|}{\textbf{Score}} &
\textbf{Acc(\%)} & \multicolumn{1}{c|}{\textbf{Score}} &
\textbf{Acc(\%)} & \multicolumn{1}{c|}{\textbf{Score}} &
\textbf{Acc(\%)} & \multicolumn{1}{c|}{\textbf{Score}} &
\textbf{Acc(\%)} & \textbf{Score} & \textbf{Ret.(\%)} \\
\midrule

\multirow{3}{*}{\textbf{Qwen2.5-VL-32B / 7B}} &
\textsc{SpecVLM}
& 34.52 & \multicolumn{1}{c|}{2.11}
& 31.12 & \multicolumn{1}{c|}{1.93}
& 28.59 & \multicolumn{1}{c|}{1.78}
& 30.85 & \multicolumn{1}{c|}{1.91}
& 31.27 & 1.93 & 100.0 \\
& +Random Relaxation (50\%)
& 31.85 & \multicolumn{1}{c|}{1.94}
& 32.63 & \multicolumn{1}{c|}{2.00}
& 28.37 & \multicolumn{1}{c|}{1.82}
& 25.13 & \multicolumn{1}{c|}{1.70}
& 29.50 & 1.87 & \textbf{95.6} \\
& +Random Relaxation (75\%)
& 31.23 & \multicolumn{1}{c|}{1.97}
& 31.87 & \multicolumn{1}{c|}{1.90}
& 26.22 & \multicolumn{1}{c|}{1.73}
& 25.63 & \multicolumn{1}{c|}{1.75}
& 28.74 & 1.84 & \textbf{93.6} \\
\midrule

\multirow{3}{*}{\textbf{Qwen2.5-VL-32B / 3B}} &
\textsc{SpecVLM}
& 34.52 & \multicolumn{1}{c|}{2.11}
& 31.12 & \multicolumn{1}{c|}{1.93}
& 28.59 & \multicolumn{1}{c|}{1.78}
& 30.85 & \multicolumn{1}{c|}{1.91}
& 31.27 & 1.93 & 100.0 \\
& +Random Relaxation (50\%)
& 31.43 & \multicolumn{1}{c|}{1.94}
& 30.90 & \multicolumn{1}{c|}{1.95}
& 15.69 & \multicolumn{1}{c|}{1.25}
& 24.43 & \multicolumn{1}{c|}{1.58}
& 25.61 & 1.68 & \textbf{84.5} \\
& +Random Relaxation (75\%)
& 28.78 & \multicolumn{1}{c|}{1.80}
& 29.46 & \multicolumn{1}{c|}{1.81}
& 20.03 & \multicolumn{1}{c|}{1.37}
& 21.07 & \multicolumn{1}{c|}{1.45}
& 24.84 & 1.61 & \textbf{81.4} \\
\midrule

\multirow{3}{*}{\textbf{Qwen2.5-VL-7B / 7B}} &
\textsc{SpecVLM}
& 30.79 & \multicolumn{1}{c|}{1.95}
& 33.44 & \multicolumn{1}{c|}{2.02}
& 25.80 & \multicolumn{1}{c|}{1.69}
& 24.77 & \multicolumn{1}{c|}{1.56}
& 28.70 & 1.81 & 100.0 \\
& +Random Relaxation (50\%)
& 29.22 & \multicolumn{1}{c|}{1.85}
& 28.37 & \multicolumn{1}{c|}{1.85}
& 26.39 & \multicolumn{1}{c|}{1.78}
& 26.15 & \multicolumn{1}{c|}{1.66}
& 27.53 & 1.79 & \textbf{97.4} \\
& +Random Relaxation (75\%)
& 27.93 & \multicolumn{1}{c|}{1.81}
& 31.30 & \multicolumn{1}{c|}{1.94}
& 25.63 & \multicolumn{1}{c|}{1.63}
& 26.96 & \multicolumn{1}{c|}{1.70}
& 27.96 & 1.77 & \textbf{97.6} \\
\bottomrule
\end{tabular}
}
\caption{Performance metric using Qwen2.5-VL on three speculative decoding settings and Video Detail Caption benchmark. Ret.(\%) refers to the performance retention ratio on average of accuracy and score, compared with autoregressive decoding.}
\label{tab:loose_sd}
\end{table*}

\begin{table}[t]
\centering
\small
\resizebox{\columnwidth}{!}{
\begin{tabular}{@{}l|l|cc@{}}
\toprule
\multirow{2}{*}{\textbf{Target / Draft Model}} &
\multirow{2}{*}{\textbf{Method}} &
\multicolumn{2}{c}{\textbf{Video Detail Caption}} \\
\cmidrule(lr){3-4}
& & \textbf{Mean Accepted Length} & \textbf{Speedup} \\
\midrule

\multirow{3}{*}{\textbf{Qwen2.5-VL-32B / 7B}} &
\textsc{SpecVLM} & 3.40 & 1.40$\times$ \\
& +Random Relaxation (50\%) & 5.55 & 2.04$\times$ \\
& +Random Relaxation (75\%) & 7.42 & 2.61$\times$ \\
\midrule

\multirow{3}{*}{\textbf{Qwen2.5-VL-32B / 3B}} &
\textsc{SpecVLM} & 2.99 & 0.97$\times$ \\
& +Random Relaxation (50\%) & 5.30 & 1.64$\times$ \\
& +Random Relaxation (75\%) & 7.33 & 2.11$\times$ \\
\midrule

\multirow{3}{*}{\textbf{Qwen2.5-VL-7B / 7B}} &
\textsc{SpecVLM} & 5.31 & 1.26$\times$ \\
& +Random Relaxation (50\%) & 7.20 & 1.61$\times$ \\
& +Random Relaxation (75\%) & 8.45 & 1.80$\times$ \\
\bottomrule
\end{tabular}
}
\caption{Efficiency metrics using Qwen2.5-VL on three speculative decoding settings and Video Detail Caption benchmark. Draft tokens per decoding step is set to 10. Decoding speedup is measured relative to autoregressive decoding.}
\label{tab:loose_sd_eff}
\end{table}

\subsection{Relaxed Speculative Decoding}
\label{ssec:exp_loosely_sd}
As discussed in~\Cref{sec:decoding}, speculative decoding for LVLMs still leaves substantial room for exploration, as many techniques in speculative decoding for LLMs~\cite{zhang2024draft,xia2025swift,shen2026doublebreakingaccelerationlimit,DBLP:journals/corr/abs-2505-16162} remain largely unexplored.
Beyond modality-aware draft models~\cite{ji2025specvlm,kong2026parallelvlm,zhang2026sparrow,shen2026mmspec}, \textit{modality-aware verification strategies} constitute another promising direction. In prevailing benchmarks for visual captioning and open-ended VQA, the importance of the output tokens may be non-uniform. Intuitively, descriptively visual tokens are more critical and thus require strict verification, whereas some prepositions, conjunctions, and other function words can be verified with relaxation~\cite{ji2026foresttreeslooselyspeculative}. This aligns with the idea that \textit{exact match} is not always needed in the recent study of relaxed speculative decoding for LLMs~\cite{bachmannjudge}.
To validate this phenomenon, we conduct a simple experiment on Qwen2.5-VL and Video Detail Caption~\cite{chaiauroracap} benchmark, using two NVIDIA H200 GPUs. The dataset is divided into four subsets, from which we sample $4\times30$ instances for evaluation.
We select the training-free speculative decoding method \textsc{SpecVLM}~\cite{ji2025specvlm} as baseline, which applies visual token pruning (90\%) to the draft model and adopts strict match verification.
As a comparison, we construct a ``random relaxation'' method, where mismatched tokens during decoding are accepted with a random probability.
The results are reported in~\Cref{tab:loose_sd_eff} and~\Cref{tab:loose_sd}.

\paragraph{Efficiency Evaluation.}
By adopting random relaxation, the verification stage of speculative decoding is directly relaxed, resulting in a largely boosted mean accepted length. Compared with \textsc{SpecVLM}, the random relaxation variants increase the decoding speedup from 1.40$\times$, 0.97$\times$, and 1.26$\times$ to 2.61$\times$, 2.11$\times$, and 1.80$\times$, respectively, effectively pushing the efficiency boundary of speculative methods.

\paragraph{Performance Evaluation.}
Remarkably, despite achieving the significant speedups mentioned above, random relaxation retains 81.4\% to 97.6\% of the original output quality across different model settings. This preservation of quality provides preliminary evidence that the output patterns of visual tasks exhibit certain sparsity, implying that many mismatches can be relaxed without severe detriment. Exploiting such output patterns, where visual entities co-occur with prepositions, conjunctions, and other function words, to perform adaptive relaxed speculative decoding remains an interesting direction for future work.




\section{Roofline Analysis Details}
\label{app:roofline_details}

\begin{figure}[t]
  \centering
  \includegraphics[width=0.48\textwidth]{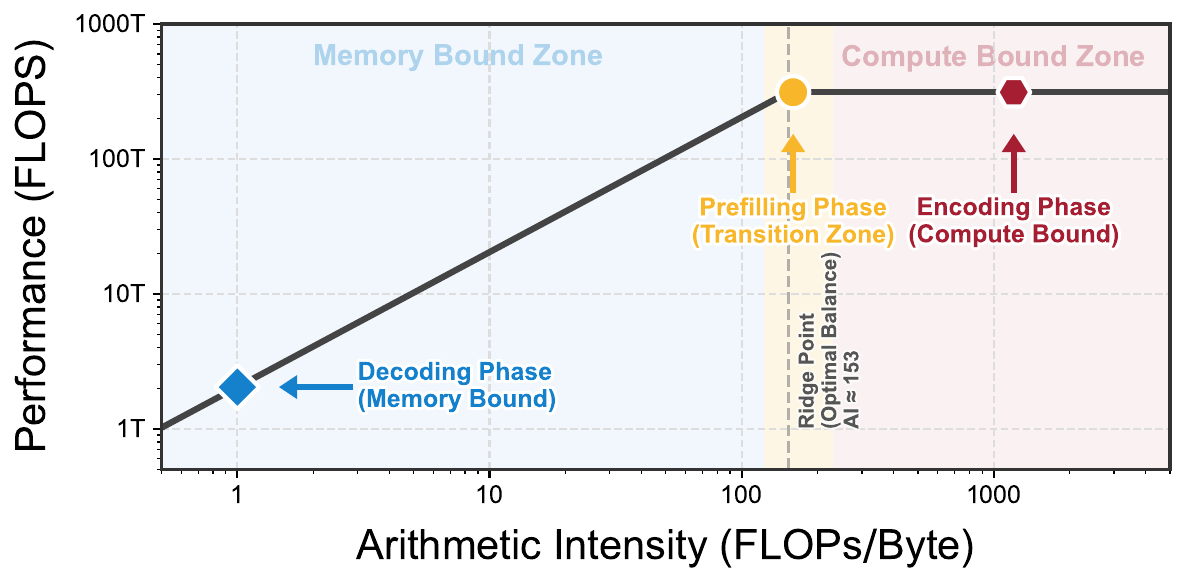}
  \vspace{-2mm} 
  \caption{
    Stage-wise bottleneck analysis of generic LVLM inference on NVIDIA A100. 
    We illustrate the operational intensity of distinct inference phases against the hardware Roofline limits. 
    The \textit{Decoding} phase is strictly memory-bound ($\mathcal{I}_a \approx 1$), constrained by bandwidth. 
    In contrast, the \textit{Visual Encoding} phase represents a compute-bound workload ($\mathcal{I}_a \approx 1200$). 
    The \textit{Prefilling} phase ($\mathcal{I}_a \approx 160$) occupies the transitional region near the hardware's ridge point ($\mathcal{I}_{ridge} \approx 153$ FLOPs$/$Byte), utilizing both compute and memory resources efficiently.
  }
  \label{fig:roofline}
  \vspace{-3mm}
\end{figure}

This section details the hardware profiling methodology and the theoretical framework underpinning the \emph{arithmetic intensity} ($\mathcal{I}$) estimations used in our Roofline analysis (see~\Cref{fig:roofline}).

\subsection{Hardware Specifications}
We employ the NVIDIA A100-SXM4-80GB GPU as the reference hardware platform. Performance limits are derived based on Half-Precision (\texttt{FP16}) tensor core operations, the standard precision for Large Vision-Language Model (LVLM) inference. The specifications are summarized in~\Cref{tab:hardware_params}.

\begin{table}[!ht]
    \centering
    \scriptsize
    \resizebox{\columnwidth}{!}{
    \begin{tabular}{@{}ll@{}}
    \toprule
    \textbf{Parameter} & \textbf{Value} \\
    \midrule
    Peak Performance ($\pi_{peak}$) & $312$ TFLOPS (\texttt{FP16} Tensor Core) \\
    Memory Bandwidth ($\beta_{mem}$) & $2,039$ GB/s (HBM2e) \\
    Ridge Point ($\mathcal{I}_{ridge}$) & $\approx 153$ \text{FLOPs$/$Byte} \\
    \bottomrule
    \end{tabular}
    }
    \caption{Hardware specifications for the NVIDIA A100 (80GB) used in the Roofline model.}
    \label{tab:hardware_params}
\end{table}

The \textit{Ridge Point} ($\mathcal{I}_{ridge}$), delineating the boundary between memory-bound and compute-bound regimes, is calculated as:
\begin{equation}
\begin{aligned}
    I_{ridge} &= \frac{\pi_{peak}}{\beta_{mem}} = \frac{312 \times 10^{12}}{2039 \times 10^9} \\ &\approx 153.0 \text{ FLOPs$/$Byte}
\end{aligned}
\end{equation}

\subsection{Workload Characterization by Stage}
We characterize the three distinct stages of modern LVLM inference by analyzing their theoretical arithmetic intensity ($\mathcal{I}_a$).

\paragraph{Decoding ($\mathcal{I}_a \approx 1.0$).} 
The decoding stage follows an autoregressive generation pattern, producing one token per step. This operation is dominated by Matrix-Vector multiplication (GEMV). For a model with parameters $\theta$, generating a single token necessitates loading the entire weight matrix to perform the computation. Under \texttt{FP16} precision (2 bytes per parameter), the intensity is derived as:
\begin{equation}
    I_{dec} \approx \frac{2 \cdot |\theta| \cdot 1 \text{ (token)}}{2 \cdot |\theta| \text{(bytes)}} = 1.0 \text{ FLOPs$/$Byte}
\end{equation}
Consequently, the decoding stage is strictly \textbf{memory-bound}, situated significantly to the left of the ridge point. Performance in this regime is solely determined by memory bandwidth utilization.

\paragraph{Prefilling ($\mathcal{I}_a \approx 160.0$).}
The prefilling stage processes the input prompt in parallel, relying on Matrix-Matrix multiplication (GEMM). Unlike decoding, the arithmetic intensity here scales with the input sequence length $(N_v+N_t)$ due to weight reuse. We visualize a representative operational point of $\mathcal{I}_a \approx 160.0$, which corresponds to moderate-to-long context lengths (e.g., $(N_v+N_t) \approx 512$). 
This value lies in the immediate vicinity of the hardware ridge point ($153.0$), indicating a mixed bottleneck regime. In this region, the workload simultaneously saturates memory bandwidth and approaches peak compute utilization, making performance highly sensitive to both data movement and arithmetic throughput optimization.

\paragraph{Encoding ($\mathcal{I}_a \approx 1200.0$).}
The encoding stage processes high-resolution image inputs via a Vision Transformer (ViT) backbone. Unlike the sparse memory access patterns in decoding, the vision encoder performs dense, highly parallel computations on image patches. We model this workload with an approximate intensity of $\mathcal{I}_a \approx 1200.0$, nearly an order of magnitude higher than the ridge point. This classifies the visual encoder as strictly compute-bound, implying that optimizations to memory bandwidth yield negligible performance gains in this stage.

\section{LLM Usage}
Large Language Models (LLMs) were used to aid in code writing and manuscript polishing.
Specifically, the usage includes refining the language, improving readability, and ensuring clarity in the paper.
It is important to note that LLMs were not involved in the ideation, research methodology, or experimental design.




\end{document}